\documentclass[10pt,twocolumn,letterpaper]{article}

\usepackage{cvpr}              % To produce the CAMERA-READY version
\definecolor{cvprblue}{rgb}{0.21,0.49,0.74}
\usepackage[pagebackref,breaklinks,colorlinks,allcolors=cvprblue]{hyperref}
\usepackage{paralist}
\usepackage{colortbl} % 提供 \rowcolor, \columncolor, \cellcolor
\usepackage{xcolor}    % 提供颜色定义，如 \definecolor，支持更多颜色格式
\usepackage{amssymb} % 提供数学符号和箭头
\usepackage{multirow} 
\usepackage{afterpage}
\usepackage{pdflscape} % 如果需要横向放置
\usepackage{pifont}
\usepackage{overpic}
\usepackage{float}

\usepackage{hyperref}
\usepackage{url}
\usepackage{marvosym}  % 通信小信封
\def\paperID{239} % *** Enter the Paper ID here
\def\confName{CVPR}
\def\confYear{2026}

\def\methodname{VQRAE}

\title{{\methodname}: Representation Quantization Autoencoders for Multimodal Understanding, Generation and Reconstruction}

\author{
Sinan Du\textsuperscript{$\rm 1,3^{* \ddagger}$},
Jiahao Guo\textsuperscript{$\rm 2,3^{* \ddagger}$},
Bo Li\textsuperscript{\rm 3 \Letter}, 
Shuhao Cui\textsuperscript{$\rm 3$},
Zhengzhuo Xu\textsuperscript{$\rm 1$},
Yifu Luo\textsuperscript{$\rm 1$}, 
Yongxian Wei\textsuperscript{$\rm 1$} \\
Kun Gai\textsuperscript{\rm 3},
Xinggang Wang\textsuperscript{\rm 2},
Kai Wu\textsuperscript{$\rm 3^{\dagger}$}, 
Chun Yuan\textsuperscript{\rm 1 \Letter}\\
[1ex]
\normalsize \textsuperscript{\rm 1}Tsinghua University,
\textsuperscript{\rm 2}Huazhong University of Science and Technology (HUST),
\textsuperscript{\rm 3}Kolors Team, Kuaishou Technology
}
\begin{document}
\maketitle

\let\thefootnote\relax\footnotetext{$*$ Equal Contibution. $\dagger$ Project Lead. \Letter Corresponding Authors.}
\let\thefootnote\relax\footnotetext{$\ddagger$ Work done during internship in Kolors Team, Kuaishou Technology.}

\begin{abstract}
Unifying multimodal understanding, generation and reconstruction representation in a single tokenizer remains a key challenge in building unified models. 
Previous research predominantly attempts to address this in a dual encoder paradigm, e.g., utilizing the separate encoders for understanding and generation respectively or balancing semantic representations and low-level features with contrastive loss. 
In this paper, we propose \textbf{{\methodname}}, a \underline{\textbf{V}}ector \underline{\textbf{Q}}uantization version of \underline{\textbf{R}}epresentation \underline{\textbf{A}}uto\underline{\textbf{E}}ncoders, which pioneers the first exploration in unified representation to produce \textbf{Continuous} semantic features for image understanding and \textbf{Discrete} tokens for visual generation within a unified tokenizer.
Specifically, we build upon pretrained vision foundation models with a symmetric ViT decoder and adopt a two-stage training strategy: first, it freezes the encoder and learns a high-dimensional semantic VQ codebook with pixel reconstruction objective; then jointly optimizes the encoder with self-distillation constraints. 
This design enables negligible semantic information for maintaining the ability of multimodal understanding, discrete tokens that are compatible for generation and fine-grained reconstruction. 
Besides, we identify the intriguing property in quantizing semantic encoders that rely on high-dimensional codebook in contrast to the previous common practice of low-dimensional codebook in image reconstruction. The semantic VQ codebook can achieve a 100\% utilization ratio at a dimension of 1536. {\methodname} presents competitive performance on several benchmarks of visual understanding, generation and reconstruction with promising scaling property in the autoregressive paradigm for its discrete merits.
\end{abstract}

\section{Introduction}
\label{sec:intro}

\begin{figure}[t]
    \centering
    \begin{overpic}[width=1.0\linewidth]{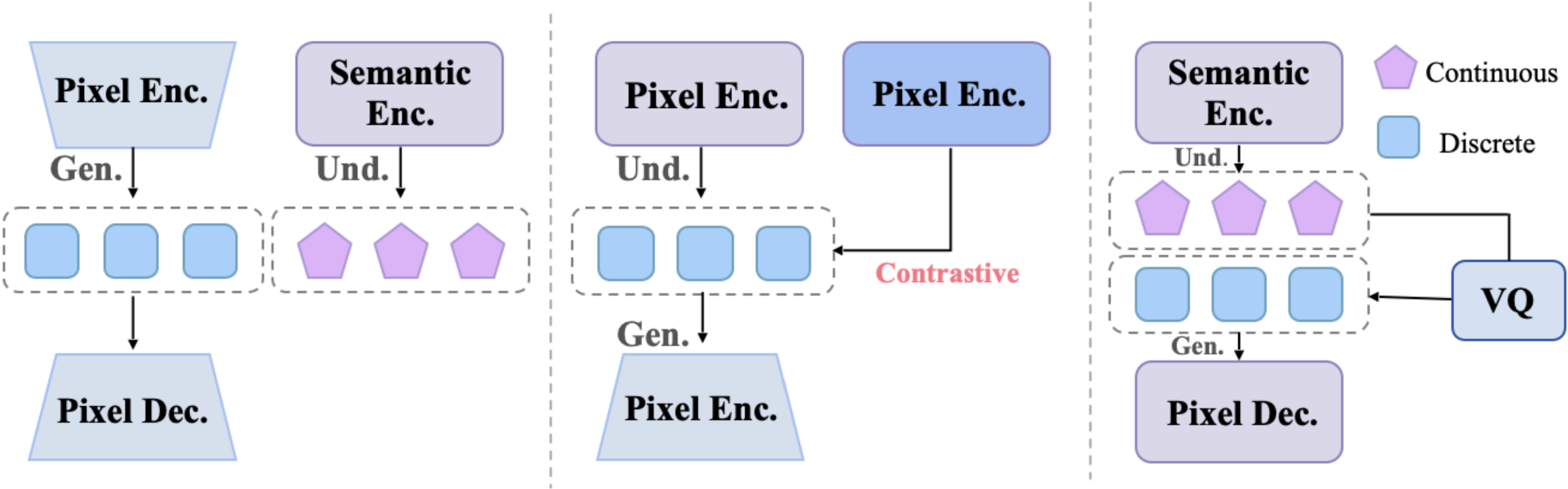}
    \put(16,-3){(a)}
    \put(49,-3){(b)}
    \put(82,-3){(c)}
    \end{overpic}
    \caption{
    Comparions of different unified tokenizers. 
    (a) Janus~\cite{janus,janus-pro} series adopt dual-encoder paradigm.
    (b) QLIP~\cite{Qlip} and UniTok~\cite{Unitok} supervise dicrete tokens with CLIP loss.
    (c) Our {\methodname} can produce continuous and discrete tokens for different tasks.
    } 
    \label{fig:intro}
    \vspace{-7pt}
\end{figure}

% \put(11,1){\tiny Understanding}    % 最小字号
% \put(11,1){\scriptsize Understanding}  % 很小字号
% \put(11,1){\footnotesize Understanding} % 小字号
% \put(11,1){\small Understanding}    % 较小字号

The advent of GPT-4o~\cite{gpt-4o} indicates the significant potential of Multimodal Large Language Models (MLLMs)~\cite{qwen3vl,internvl3.5} to unify visual understanding and generation within a single augoregressive architecture~\cite{gemini,emu3.5,Fluid,Transfusion,show-o}. These unified models deliver precise response in multimodal interaction~\cite{emu3.5,gpt-4o,gemini}, emergent in-context reasoning properties~\cite{bagel,Ming-UniVision} and synergistic benefit across tasks~\cite{reca,yan2025can}. However, a fundamental dilemma remains in selecting visual tokenizers to obtain appropriate representations to achieve a trade-off between \textit{understanding, generation, and reconstruction tasks}.

Discrete tokenizers~\cite{vqvae,vqgan} were widely adopted in early unified models~\cite{chameleon,show-o,emu3,Lumina-mgpt} due to their compatibility with the next token prediction (NTP) paradigm, scalability potential, and training efficiency enabled by highly optimized AI infrastructure. 
However, discrete tokenizers trained with reconstruction objective (pixel-level) tend to produce fine-grained details, which conflict with the semantic-level features required for visual understanding tasks such as CLIP~\cite{clip,siglip,siglip2}, leading to performance degradation.
Dual encoder methods attempt to address this by utilizing separate encoders~\cite{janus,janus-pro,janusflow} for multimodal understanding and visual generation, or balancing semantic representations and low-level features with contrastive loss. These dual encoder approaches increase model complexity~\cite{tokenflow,muse-vl}, hinder deeper interaction between representations~\cite{janus,janus-pro,janusflow}, and demand an immense batch size to balance loss conflicts effectively~\cite{Qlip,Vila-u,Unitok}.
\textit{Can we design a \textbf{unified} tokenizer to produce \textbf{continuous semantic} features and \textbf{discrete fine-grained} tokens \textbf{simultaneously}?}

% 在需要插入图片的地方
\afterpage{
    \begin{figure*}[p]
        \centering
        \begin{overpic}[width=0.96\textwidth, height=0.96\textheight]{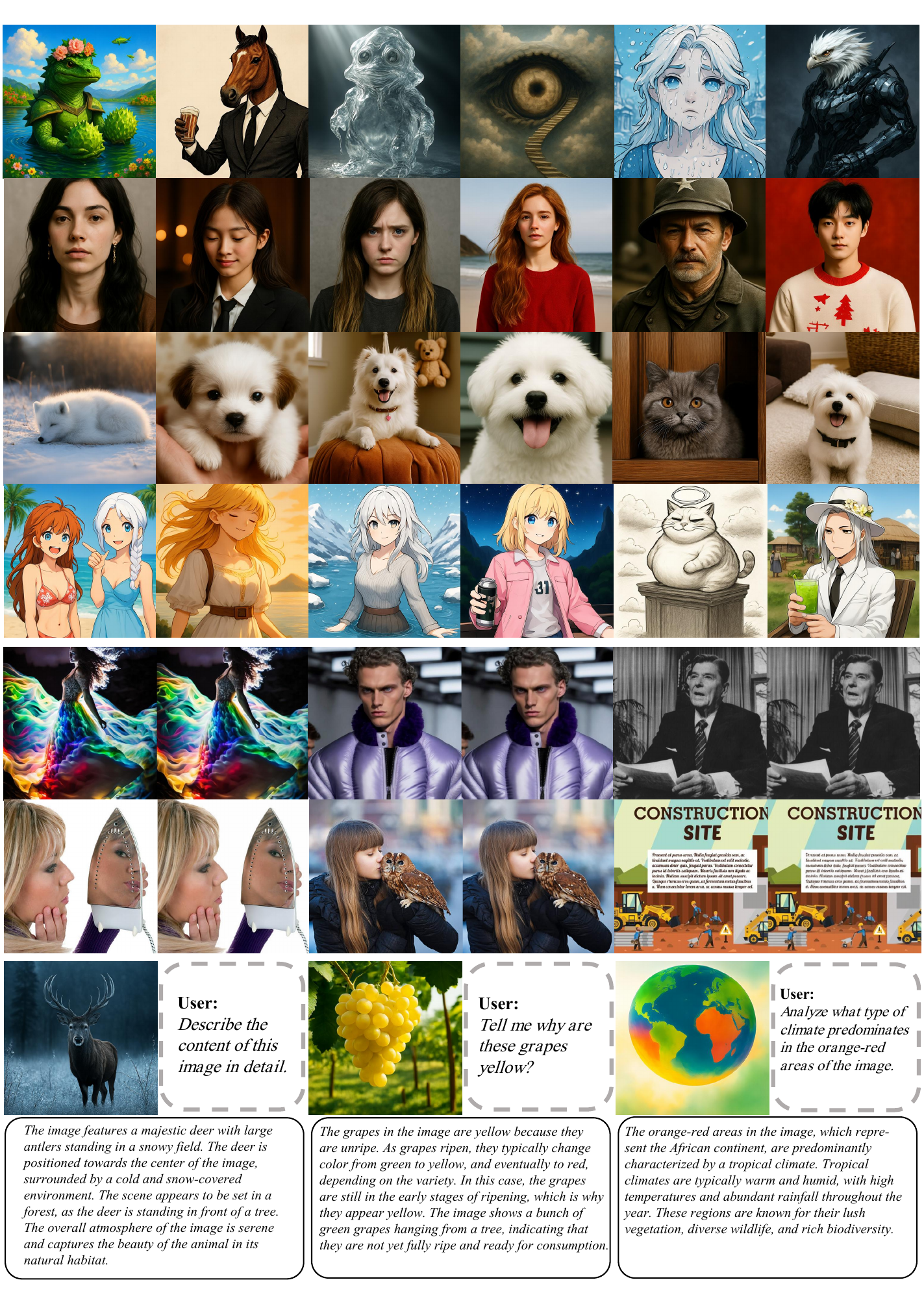}
            \put(-2,71){\rotatebox{90}{\textbf{Generation}}}
            \put(-2,33){\rotatebox{90}{\textbf{Reconstruction}}}
            \put(-2,8){\rotatebox{90}{\textbf{Understanding}}}
        \end{overpic}
        \caption{
            Showcase of the visual understanding, generation and reconstruction ability of our \textbf{{\methodname}} model.
        } 
        \label{fig:teaser}
    \end{figure*}
    \clearpage % 确保在新的一页继续文字
}

% To address this, hybrid methods~\cite{tokenflow,muse-vl,uniflow,Unitok,blip3-o,emu-2,Qlip,DualViTok,Semhitok,Dualtoken} aim to unify representations within a single tokenizer. Diffusion-based tokenizers~\cite{emu-2,blip3-o} utilize a frozen semantic encoder to maintain performance in visual understanding and latent diffusion model to recover pixel; these methods fail to achieve faithful reconstruction and are incompatible with NTP loss. TokenFlow~\cite{tokenflow} and MUSE-VL~\cite{muse-vl} adopt a dual encoder training paradigm and decouple semantic and pixel-level feature learning while maintaining their alignment via a shared mapping mechanism. QLIP~\cite{Qlip}, VILA-U~\cite{Vila-u} and UniTok~\cite{Unitok} supervise the latent features from vision foundation models~\cite{clip,siglip} with contrastive learning loss~\cite{clip}, which demand an immense batch size to train effectively and balance loss conflicts. \textit{Can we design a \textbf{unified} tokenizer with \textbf{semantic} feature, \textbf{discrete} merits and \textbf{reconstruction} ability?}

Inspired by the practice of RAE~\cite{rae} in generative modeling, which replaces the VAE~\cite{vae} with pretrained vision encoders~\cite{dino,siglip} paired with trained decoders and demonstrates the structured semantic space benefits convergence of diffusion transformers~\cite{dit}. 
We propose \textbf{{\methodname}}, a vector quantization version of RAE~\cite{rae}, which pioneers the unified tokenizer to produce continuous features for visual understanding and discrete tokens for generation and reconstruction, effectively reducing the complexity of dual encoder design and reliance on convolutional pixel encoders.
{\methodname} comprises pretrained vision foundation models (VFMs), a high-dimensional VQ codebook and a symmetric ViT decoder. We adopt a \textit{two-stage} training strategy: in the first stage, built upon the frozen VFMs, we jointly optimize the codebook to learn discrete semantic representation and the ViT decoder with pixel reconstruction objective; the second stage unfreezes the VFMs with self-distillation loss to maintain semantic features and reconstruction loss to supplement fine-grained details. Besides, to our knowledge, we are the first work to train a 100\% utilization ratio semantic codebook for reconstruction at a high dimension (\textit{e.g.}, 1536), while previous studies mainly at a low dimension (\textit{e.g.}, 8-256)~\cite{llamagen,vqgan,ibq,vqgan-lc}. 

Our key contributions are summarized as follows:
\begin{compactitem}
    \item We propose {\methodname}, a VQ version of RAE, which is the first work to train a 100\% utilization ratio semantic codebook at a high dimension for reconstruction.
    \item {\methodname} pioneers the unified tokenizer without convolution blocks, producing continuous features and discrete tokens simultaneously for understanding, generation and reconstruction under the AR-only paradigm.
    \item Extensive experiments on downstream benchmarks present competitive performance and scaling promise, opening new avenues for exploring unified models.
\end{compactitem}

\section{Related Work}
\label{sec:related}

\subsection{Visual Tokenizer for Generation}
Visual tokenizer plays a crucial role in encoding raw pixels into compact latent representations for generative modeling. Inspired by the success of next token prediction (NTP) paradigm, vector quantization (VQ) tokenizer~\cite{vqvae,vqgan,vqgan-lc,simvq,ibq} is widely explored for its discrete property and compatibility with autoregressive and masked generative models~\cite{llamagen,var,maskgit,magvit,dalle}. However, discrete methods such as Chameleon~\cite{chameleon}, EMU-3~\cite{emu3}, and Show-o~\cite{show-o} suffer performance degradation in visual understanding tasks due to quantization errors~\cite{Lumina-mgpt,simplear,pure}. In addition, the VQ tokenizer optimized with the reconstruction objective tends to provide pixel-level tokens for fine-grained features, which introduce significant alignment costs with LLMs.

\subsection{Unified Tokenizer}
To address the representation dilemma (continuous \textit{vs.} discrete, pixel \textit{vs.} semantic) above, Janus series~\cite{janus,janus-pro,janusflow} propose to disentangle representations for understanding and generation with separate visual encoder. TokenFlow~\cite{tokenflow} and MUSE-VL~\cite{muse-vl} adopt a dual encoder training paradigm and decouple semantic and pixel-level feature learning while maintaining their alignment via a shared mapping mechanism. QLIP~\cite{Qlip}, VILA-U~\cite{Vila-u} and UniTok~\cite{Unitok} supervise the latent features from vision foundation models~\cite{clip,siglip} with contrastive learning loss~\cite{clip}, which demand an immense batch size to train effectively and balance loss conflicts. However, this dual encoder paradigm leads to training inefficiency and hinders the deeper alignment and interaction between different representations, which is critical for unified models~\cite{bagel,emu3.5,metaquery,unipic,unipic-2,unilip,Ming-UniVision,OmniGen2,reca,Nextstep-1,uniworld}. Diffusion-based tokenizers~\cite{blip3-o,emu-2,uniflow} typically employ continuous representations for reconstruction, which are difficult to converge in the autoregressive paradigms due to the high dimensional CLIP feature~\cite{mar,clip}. Tar~\cite{tar} and X-Omni~\cite{x-omni} propose VQ tokenizers trained with semantic supervision but discarded reconstruction capabilities, thus precluding their nature as autoencoders. In contrast, {\methodname} achieves a superior trade-off between visual understanding and reconstruction and keeps its discrete merits for autoregressive generation.

\section{Method}
\label{sec:method}

Our \textbf{\methodname} is a vector quantization version of representation autoencoder
designed to achieve unified multimodal understanding, generation and reconstruction. 
In Sec.~\ref{sec:pilot}, we conduct pilot experiments to clarify the motivation behind our design and highlight the distinctive features of {~\methodname} compared to other approaches.
In Sec.~\ref{sec:unified}, we provide a detailed description of the {\methodname} architecture, which adopts a symmetric ViT encoder-decoder structure and a two-stage training strategy to optimize the discrete codebook.
In Sec.~\ref{sec:codebook}, we train a high-dimensional codebook when quantizing VFMs, which contrasts with previous findings that low-dimensional codebook is crucial for reconstruction and generation~\cite{vqgan,llamagen,ibq}.
In Sec.~\ref{sec:und} and ~\ref{sec:gen}, we train understanding and generation with {\methodname}.

\subsection{Pilot Experiments}
\label{sec:pilot}
\begin{table}[h!]
\centering
\scalebox{0.49}{
    \begin{tabular}{ccccc|ccc|ccc}
    \toprule
    \textbf{\# Exp.} & \textbf{Method} & \textbf{Und.} & \textbf{Gen.} & \textbf{Type} & \textbf{rFID$\downarrow$} & \textbf{PSNR$\uparrow$} & \textbf{SSIM$\uparrow$} & \textbf{MME-P$\uparrow$} & \textbf{SEED$\uparrow$} & \textbf{TQA$\uparrow$} \\
    \midrule
    1 & VQGAN\textsuperscript{\dag}~\cite{vqgan} & D & D & single & 4.98 & 20.00 & 0.629 & 756.1 & 38.2 & 46.8 \\
    2 & VQKD\textsuperscript{\dag}~\cite{vqkd} & D & - & single & - & - & - & 1252.4 & 57.8 & 48.2 \\
    3 & RAE\textsuperscript{\dag}~\cite{rae} & C & C & single & \textbf{0.49} & 19.23 & 0.620 & \textbf{1544.3} & \textbf{70.0} & \textbf{61.7} \\
    4 & TokenFlow\textsuperscript{\dag}~\cite{tokenflow} & D & D & dual & 1.37 & 21.41 & 0.690 & 1365.4 & 62.6 & 54.1 \\ % TokenFlow-L
    5 & Janus~\cite{janus} & C & D & dual & 2.19 & 20.79 & 0.675 & 1338.0 & 63.7 & - \\
    \midrule
    \rowcolor{cyan! 10} 6 & \textbf{{\methodname}\textsuperscript{\dag} (ours)} & C & D & single & 1.31 & \textbf{22.23} & \textbf{0.762} & 1543.3 & \textbf{70.0} & \textbf{61.7}\\
    \bottomrule
    \end{tabular}
}

\caption{
Comparisons of various methods in multimodal understanding and reconstruction. ``Und.'' and ``Gen.'' refer to Understanding and Generation. 
``C'' and ``D'' indicate Continuous and Discrete representations used for specific tasks.
\dag\ denotes training on LLaVA-v1.5~\cite{llava-1.5} setting. 
Reconstruction quality is evaluated on the 256 $\times$ 256 ImageNet 50k validation set. MME-P: MME-Perception~\cite{mme}; SEED: SEEDBench-Img~\cite{seedb}; TQA: TextVQA~\cite{textvqa}; TokenFlow: TokenFlow-L-13B~\cite{tokenflow}.
}
\vspace{-5pt}
\label{tab:pilot}
\end{table}

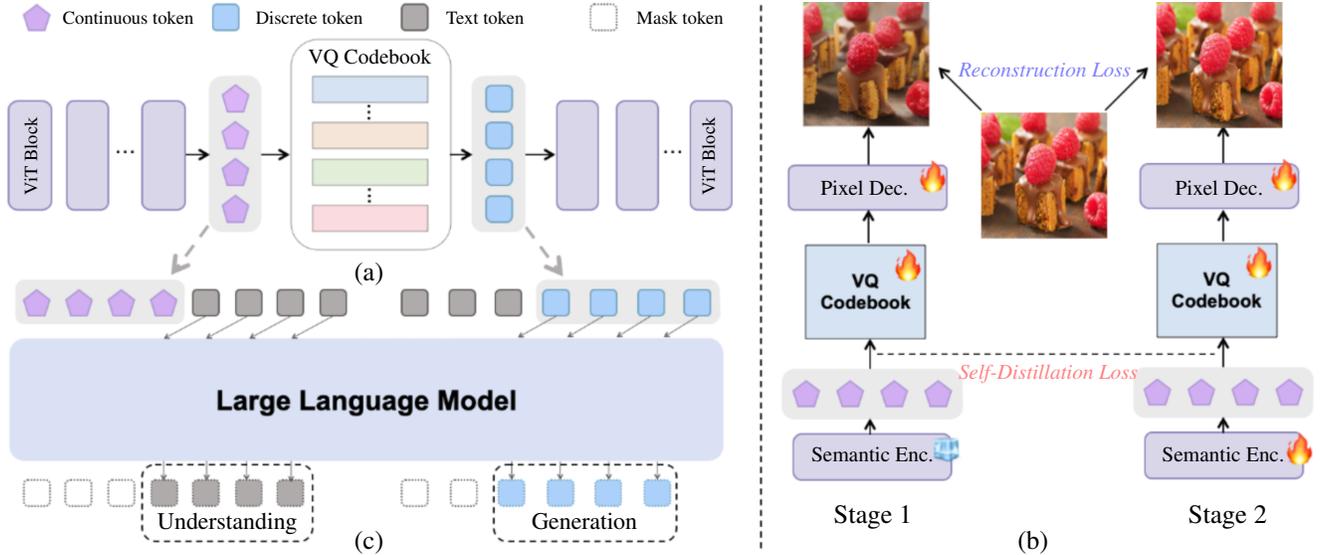
\begin{figure*}[t]
    \centering
    \begin{overpic}[width=1.0\textwidth]{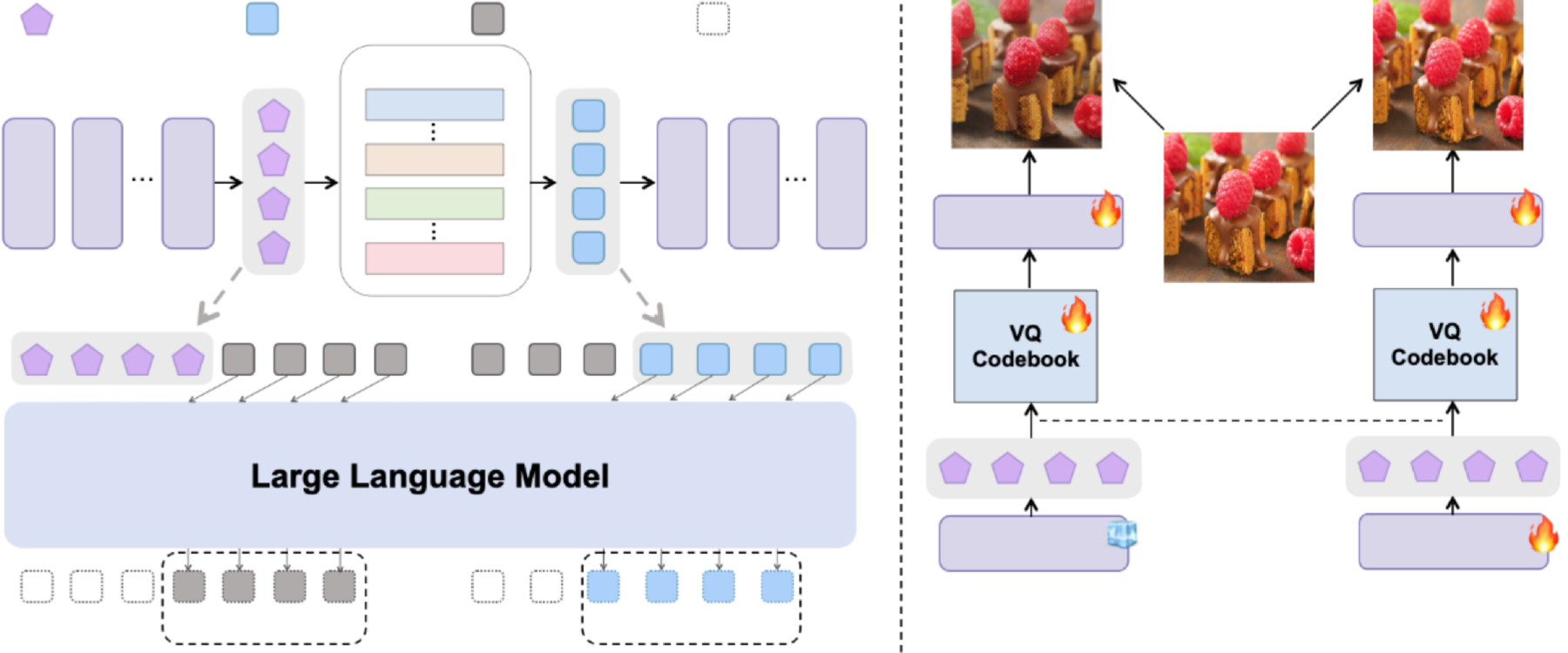}
    % \put(-6,33){(mark)}
    % \put(25,19){(a)}
    % \put(4.3,37.5){\scriptsize Continuous token}
    % \put(18,37.5){\scriptsize Discrete token}
    % \put(31.5,37.5){\scriptsize Text token}
    % \put(45,37.5){\scriptsize Mask token}
    % \put(2,25){\rotatebox{90}{\scriptsize ViT Block}}
    % \put(22,34.5){\footnotesize VQ Codebook}
    % \put(51,25){\rotatebox{90}{\scriptsize ViT Block}}
    
    % \put(76,0){(b)}
    % \put(60,2){Stage 1}
    % \put(90,2){Stage 2}
    % \put(58.3,5.7){\footnotesize Semantic Enc.}
    % \put(59,25){\footnotesize Pixel Dec.}
    % \put(71,12){\footnotesize \textit{\textcolor{red!50}{Self-Distillation Loss}}}
    % \put(71,34){\footnotesize \textit{\textcolor{blue!50}{Reconstruction Loss}}}
    % \put(88,5.7){\footnotesize Semantic Enc.}
    % \put(89,25){\footnotesize Pixel Dec.}
    
    % \put(25,0){(c)}
    % \put(11,1){\small Understanding}
    % \put(38,1){\small Generation}
    \put(26.5,20.5){(a)}
    \put(4.3,40){\scriptsize Continuous token}
    \put(19,40){\scriptsize Discrete token}
    \put(33.5,40){\scriptsize Text token}
    \put(48,40){\scriptsize Mask token}
    \put(1.4,27){\rotatebox{90}{\scriptsize ViT Block}}
    \put(23,37){\footnotesize VQ Codebook}
    \put(53,27){\rotatebox{90}{\scriptsize ViT Block}}
    
    \put(77,0){(b)}
    \put(63,2){Stage 1}
    \put(90,2){Stage 2}
    \put(61.3,6.7){\footnotesize Semantic Enc.}
    \put(62,27){\footnotesize Pixel Dec.}
    \put(72.5,13){\footnotesize \textit{\textcolor{red!50}{Self-Distillation Loss}}}
    \put(72.5,36){\footnotesize \textit{\textcolor{blue!50}{Reconstruction Loss}}}
    \put(88,6.7){\footnotesize Semantic Enc.}
    \put(89,27){\footnotesize Pixel Dec.}
    
    \put(26.5,0){(c)}
    \put(11.5,1.5){\small Understanding}
    \put(40,1.5){\small Generation}
    \end{overpic}
    \caption{
    Illustration of our unified tokenizer \textbf{{\methodname}}. 
    (a) Our {\methodname} is built on pretrained VFMs (\textit{e.g.}, SigLIP2~\cite{siglip2}), which can simultaneously produce continous semantic features for multimodal understanding tasks and discrete tokens for visual generation and reconstruction tasks.
    (b) Training pipeline of {\methodname}. We adopt a two-stage training paradigm. In the first stage, the pretrained semantic encoder remains frozen, while a high-dimensional vector quantization codebook and a pixel decoder are trained using an image reconstruction loss. In the second stage, the encoder, codebook, and decoder are jointly optimized to achieve fine-grained reconstruction. Additionally, the encoder outputs are regularized via a self-distillation loss to maintain semantic understanding performance.
    (c) {\methodname} achieves a superior trade-off with the unified encoder in the autoregressive style.
    } 
    \label{fig:method}
    \vspace{-6pt}
\end{figure*}

% \put(11,1){\tiny Understanding}    % 最小字号
% \put(11,1){\scriptsize Understanding}  % 很小字号
% \put(11,1){\footnotesize Understanding} % 小字号
% \put(11,1){\small Understanding}    % 较小字号

To further elucidate the design rationale behind {\methodname} for achieving task-level trade-off within a unified tokenizer, we conduct pilot experiments utilizing the LLaVA-1.5~\cite{llava-1.5} settings for image understanding and the ImageNet dataset for reconstruction. 
In Tab.~\ref{tab:pilot} Exp. 1-2, the training objective of pixel reconstruction yields fine-grained features that underperform in multimodal understanding tasks; 
Although distilling VFMs~\cite{vqkd} for discrete tokens can narrow this performance gap, the resulting representations still lag behind those produced by continuous tokenizers, which are attributed to the quantization errors in discretization.
TokenFlow~\cite{tokenflow} addresses this with semantic and pixel encoder sharing a mapping network, while Janus~\cite{janus} directly utilizes separate encoders.
% They both adopt dual encoder paradigm.
In contrast, {\methodname}, is capable of generating both continuous features for visual understanding and discrete tokens for generation and reconstruction within a single tokenizer. 
This differs from the dual encoder architecture employed by TokenFlow and Janus, and enables a more favorable trade-off as presented in Exp. 4-6.

\subsection{A Unified Tokenizer: {\methodname}}
\label{sec:unified}
As illustrated in Fig.\ref{fig:method}a, our {\methodname} is composed of a unified tokenizer for both low-level and semantic encoding with well-pretrained VFMs, a high-dimensional semantic VQ codebook with 100\% utilization ratio, and a symmetric ViT decoder for pixel reconstruction.

\textbf{\textit{VFMs as Unified Encoder.}}
Unlike prior works in dual encoder paradigm~\cite{tokenflow,janus,muse-vl} that employ a semantic encoder (ViT-based) and a pixel encoder (CNN-based) for separate representations, this approach introduces additional model complexity and training overhead, while also leading to a disparity in representation interaction. 
We adopt the pretrained VFMs (\textit{e.g.,} CLIP~\cite{clip,siglip2,internvl3}) as unified encoder $E$ for encoding visual information. 
We find that the continuous features generated by the frozen semantic encoder can be directly utilized for image reconstruction, albeit with some loss of detail—such as in color and texture as evidenced in ~\cite{unilip,rae}. 
However, slight fine-tuning of the encoder can refine the representations to recover these missing details, while causing almost no degradation but even stronger in semantic understanding. 
Given an input image $X \in \mathbb{R}^{h\times w\times 3}$ and the unified encoder $E$ with patch size $p$ and hidden size $d$, we obtain latent features $Z_I \in \mathbb{R}^{\frac{hw}{p^2}\times d}$. 
These intermediate features are used for semantic quantization and multimodal understanding task.

\textbf{\textit{High Dimensional VQ.}} 
Vector Quantization~\cite{vqgan,vqvae} is a well-studied technique used to translate continuous representations into a set of discrete tokens. 
Unlike previous discrete unified tokenizers~\cite{tokenflow,muse-vl} built upon pixel features, we only exploit the semantic features from VFMs for quantization. 
Specifically, we utilize SimVQ~\cite{simvq} method with initialized VQ codebook $C \in \mathbb{R}^{k\times e} = \{c^i\}|_{i=1}^k$ and learnable projection matrix $W\in \mathbb{R}^{e\times e} = \{w^i\}|_{i=1}^e$ for quantization, where $n$ is the codebook size and $e$ denotes the codebook dimension. 
Semantic features $Z_I$ from VFMs are projected to $\hat{Z_c}\in\mathbb{R}^{\frac{hw}{p^2}\times e}$ and then quantized vectors $Z_q\in\mathbb{R}^{\frac{hw}{p^2}\times e}$ are selected from codebook $C$ according to the $l_2$-norm distances:
\begin{equation}
Z_q=lookup\big(\underset{i}{argmin}||\hat{Z}_c - c^iw^i||\big), \text{for } i = 1, \ldots, k
\end{equation}
We highlight that the codebook in our {\methodname} performs effectively in high-dimensional formulation, with its dimensionality necessarily matching at least that of the VFMs encoder $E$. 
This observation contrasts with previous studies, which suggest that codebooks for training reconstruction objectives should operate in low-dimensional spaces~\cite{llamagen,vqgan,ibq}. 
This discrepancy will be presented in our ablations.

\textbf{\textit{Symmetric Decoder.}} 
We replace the previous CNN-like pixel decoder~\cite{vqgan,vae,unilip} with a ViT-based decoder that mirrors the encoder structure, thereby mapping latent features back to pixel space as in RAE~\cite{rae}, while our approach focuses on reconstructing images from discrete tokens. 
Specifically, quantized vectors $Z_q$ are projected to the bottleneck features $Z_{bot}\in \mathbb{R}^{\frac{hw}{p^2}\times d}$ to align with dimension of the symmetric decoder $D$.
We simply set the patch size of the decoder as 1 and project the decoded features $D(Z_{bot})$ from decoder to $X'\in \mathbb{R}^{\frac{hq}{p}\times\frac{wq'}{p}\times 3}$, where $q$ and $q'$ are hyperparameters to adjust the resolution of the reconstructed images. We set $q=q'=p$ to keep the resolution constant.

\textbf{\textit{Two-Stage Training.}} 
In the original RAE~\cite{rae} paper, VFMs encoder keeps frozen to maintain the structure of semantic features. 
However, VFMs are not meant to be optimized for fine-grained reconstruction and may cause blur in this setting. 
Our pilot experiments in Sec.~\ref{sec:pilot} validate that appropriate finetuning can strengthen the reconstruction ability and VFMs encoder are robust to maintain the performance of visual understanding. 
As detailed in Fig.~\ref{fig:method}b, in the first stage, we freeze the VFMs encoder $E$ and jointly optimize the VQ codebook $C$ and decoder $D$ with the pixel reconstruction and vector quantization objectives:
\begin{equation}
\mathcal{L}_{\text{rec}} = {\ell}_2(X, X') + \mathcal{L}_{\text{P}}(X, X') + \lambda_{\text{G}} \mathcal{L}_{\text{G}}(X')
\end{equation}
\begin{equation}
\mathcal{L}_{\text{quant}} = ||sg(C)-Z_q||_2^2 + \beta \cdot ||Z_q-sg(C)||_2^2
\end{equation}
\begin{equation}
\mathcal{L}_{\text{stage1}} = \mathcal{L}_{\text{rec}} + \mathcal{L}_{\text{quant}}
\end{equation}
where ${\ell}_2$ represents pixel-wise reconstruction loss, $\mathcal{L}_{\text{P}}(\cdot)$ denotes perceptual loss using LPIPS, $\text{sg}[\cdot]$ denotes the stop-gradient operation, and $\mathcal{L}_{\text{G}}(\cdot)$ represents adversarial loss with $\lambda_{\text{G}}$ as its weight coefficient. $\beta$ is set to $0.25$ by default.

As depicted in Fig.~\ref{fig:method}b, in the second stage, we unfreeze the VFMs encoder with self-distillation loss constraints to augment reconstruction quality and maintain the performance of multimodal understanding.
Note that, we directly supervise the continuous features $Z_I$ without quantization errors as in ~\cite{vqkd,tar}. The loss function in this stage is:
\begin{equation}
\mathcal{L}_{\text{distill}}=||Z_I - T(X)||_2^2
\end{equation}
\begin{equation}
\mathcal{L}_{\text{stage2}} = \mathcal{L}_{\text{rec}} + \mathcal{L}_{\text{quant}} + \lambda_{d}\mathcal{L}_{\text{distill}}
\end{equation}
where $T$ denotes the teacher model initialized from $E$ and keeps frozen, while encoder, VQ codebook and decoder are jointly optimized.
$\lambda_{d}$ is the coeffienct of distillation loss.

\subsection{Semantic VQ with 100\% Utilization Ratio.}
\label{sec:codebook}
Previous VQ methods (\textit{e.g.}, VQVAE~\cite{vqvae} and VQGAN~\cite{vqgan,llamagen}) typically train a discrete codebook at low dimensional entries (\textit{e.g.}, 8-256) with the pixel reconstruction objective with CNN extracted features.
As discussed in ~\cite{vqgan,llamagen,ibq}, it remains a fundamental challenge to train a high dimensional codebook without collapse risks and with a high utilization ratio. 
RAE~\cite{rae} illustrates that continuous high dimensional latent space is more structured and can facilitate the convergence of state-of-the-art generative models~\cite{dit,sit,repa}.
To our knowledge, our {\methodname} represents the first attempt to successfully train a high dimensional codebook at a 100\% utilization ratio (4k-16k entries). 
Contrary to the findings in previous CNN-based VQ codebook practices~\cite{vqgan,llamagen,ibq}, our empirical results indicate that a semantic codebook generally requires a larger dimension; otherwise, it may lead to non-convergence in reconstruction training and codebook collapse.

\subsection{Multimodal Understanding with \methodname}
\label{sec:und}
Different from previous works~\cite{tar,vqkd} which utilize self-distillation loss to supervise the discrete tokens, our {\methodname} can produce intermediate continuous features for image understanding without quantization errors.
Furthermore, since our {\methodname} is built upon pretrained VFMs, it can be seamlessly integrated into existing MLLMs, significantly reducing training overhead and the need for separate evaluation of the performance of the tokenizer. 
This approach eliminates the previous requirement to complete tokenizer and MLLM training before assessing the visual comprehension capabilities of the model. 
For instance, by employing the VFMs from existing MLLMs as the encoder, the tokenizer obtained through our two-stage training can be directly incorporated into MLLMs without requiring additional pretraining or supervised fine-tuning. 
\textit{\textbf{Consequently, the MLLMs presented in this paper have not been specifically trained for the proposed unified tokenizer.}}

\begin{figure}[t]
    \centering
    \begin{overpic}[width=1.0\linewidth]{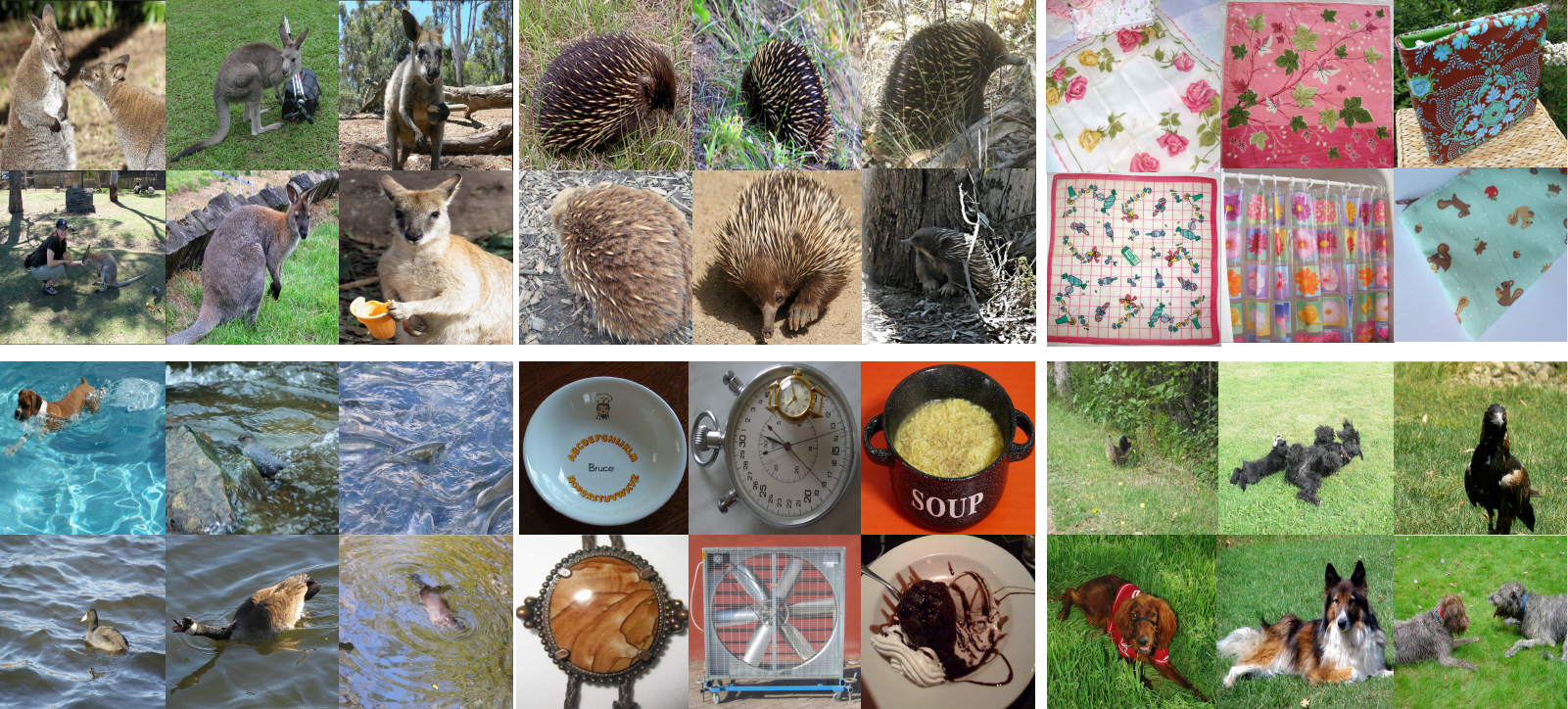}
    \put(-6,33){(a)}
    \put(-6,10){(b)}
    \end{overpic}
    \caption{
    We perform K-means clustering on the ImageNet-1K validation set using continuous features and discrete tokens.
    The visualization illustrates images grouped by (a) continuous features and (b) discrete tokens, \textbf{both derived from our {\methodname}}. 
    {\methodname} is capable of producing \textit{discriminative} features for multimodal understanding and discrete visual tokens for \textit{fine-grained} reconstruction and generation simultaneously within a unified tokenizer.
    It indicates the redundancy in the dual-encoder paradigm.
    } 
    \label{fig:cluster}
    \vspace{-15pt}
\end{figure}

\subsection{Visual Generation with \methodname}
\label{sec:gen}
Although MAR~\cite{mar,Nextstep-1} methods demonstrate the ability of continuous autoregressive for generative modeling, our discrete VQ codebook is more elegant and compatible with highly optimized AI infrastructure for training acceleration. 
We leverage the text tokenizer for text encoding and {\methodname} for image encoding.
Building on the backbone of Qwen3~\cite{qwen3}, we expand the vocabulary size for visual tokens and train the LLMs with NTP loss exclusively on visual tokens.
We highlight the property of our disentangled representations with a unified tokenizer, as visualized in Fig.~\ref{fig:cluster}, the semantic features tend to cluster similar objects and animals, while fine-grained features cluster same textures. We use the {\methodname}-InternViT version with codebook size of 16k and dimension of 1536.
% It indicates that {\methodname} can produce discriminative features at the semantic level for understanding and fine-grained features for visual generation in a \textbf{\textit{unified}} tokenizer.

\section{Experiments}
\label{sec:experiments}

\subsection{Experimental Setups}
\label{sec:setting}

\textbf{Datasets.}
{\methodname} is pretrained on BLIP3-o~\cite{blip3-o} open-sourced data, which consists of 27M samples recaptioned by Qwen2.5-VL-7B~\cite{Qwen2.5-VL}, 5M samples from CC12M~\cite{cc12m}, and 4M synthesized images from JourneyDB~\cite{journeydb}. 
For image understanding, we follow the setups in LLaVA-1.5~\cite{llava-1.5}, which uses LLaVA-Pretrain-595K and LLaVA-v1.5-mix-665K for pretraining and SFT. 
For visual generation, we train it on BLIP3-o~\cite{blip3-o} data and additional 80M high quality images. 
We conduct ablation studies on VQ codebook on ImageNet-1K with 20 epochs training for efficiency.

% siglip2-so400m-p16-256px-vicuna-7b, siglip2-so400m-p16-512px-vicuna-7b
% siglip2-so400m-p16-256px-vicuna-13b, siglip2-so400m-p16-512px-vicuna-13b
% InternViT-300m

\hspace{-0.4cm}\textbf{Implementation Details.}
We implement {\methodname} with SigLIP2-so400m-p16-256px~\cite{siglip2}, SigLIP2-so400m-p16-512px~\cite{siglip2} and InternViT-300M-448px~\cite{internvl3} as unified encoders.
Detailed hyperparameters are provided in the Appendix. 
For image understanding, we employ Vicuna-v1.5~\cite{vicuna} and Qwen2.5-7B~\cite{qwen2.5} as LLM backbone. 
Note that, we did not conduct training specifically for our {\methodname} as clarified in Sec.~\ref{sec:und}. 
For visual generation, we use Qwen3-0.6B~\cite{qwen3} for efficient training.

\hspace{-0.4cm}\textbf{Evaluation Metrics.}
We assess reconstruction quality using rFID, PSNR, and SSIM on the ImageNet-1K validation set. 
For multimodal understanding, we evaluate on MME-Perception~\cite{mme}, GQA~\cite{gqa}, POPE~\cite{pope}, MMBench-en~\cite{mmbench}, SEEDBench-Img~\cite{seedb}, MMMU~\cite{mmmu}, TextVQA~\cite{textvqa} and AI2D~\cite{ai2d}.
For image generation, we evaluate on GenEval~\cite{geneval} and DPG-Bench~\cite{dpgbench}.

\begin{table}[h!]
\centering
\scalebox{0.85}{
    \begin{tabular}{c|c|ccc}
    \toprule
    \textbf{Method} & \textbf{Ratio} & \textbf{rFID$\downarrow$} & \textbf{PSNR$\uparrow$} & \textbf{SSIM$\uparrow$} \\
    \midrule
    \multicolumn{5}{c}{\textit{Generative Only Tokenizer}} \\
    \midrule
    VQGAN~\cite{vqgan} & 16 & 4.98 & 20.00 & 0.629 \\
    LlamaGen~\cite{llamagen} & 16 & 2.19 & 20.79 & 0.675 \\
    VAR~\cite{var} & 16 & 1.00 & 22.63 & 0.755 \\
    Open-MAGVIT2~\cite{open-magvit2} & 16 & 1.67 & 22.70 & 0.640 \\
    RAE~\cite{rae} & 16 & \textbf{0.49} & 19.23 & 0.620 \\
    \midrule
    \multicolumn{5}{c}{\textit{Unified Tokenizer}} \\
    \midrule
    Show-o~\cite{show-o} & 16 & 3.50 & 21.34 & 0.590 \\
    TokenFlow~\cite{tokenflow} & 16 & 1.37 & 21.41 & 0.690 \\
    DualViTok~\cite{DualViTok} & 16 & 1.37 & 22.53 & 0.740 \\
    MUSE-VL~\cite{muse-vl} & 16 & 2.26 & 20.14 & 0.646 \\
    \rowcolor{cyan! 10} \textbf{{\methodname} (SigLIP2)} & 16 & 1.31 & 22.23 & 0.762 \\
    \rowcolor{cyan! 10} \textbf{{\methodname} (InternViT)} & 14 & 1.39 & \textbf{22.88} & \textbf{0.784} \\
    \bottomrule
    \end{tabular}
}

\caption{
Comparisons of reconstruction quality on the 256 $\times$ 256 ImageNet 50k validation set.
Ratio: downsample ratio.
}
\label{tab:tokenizer}
\vspace{-8pt}
\end{table}
\subsection{Unified Visual Tokenizers}
\label{sec:tokenizers}

\begin{figure}[t]
    \centering
    \begin{overpic}[width=1.0\linewidth]{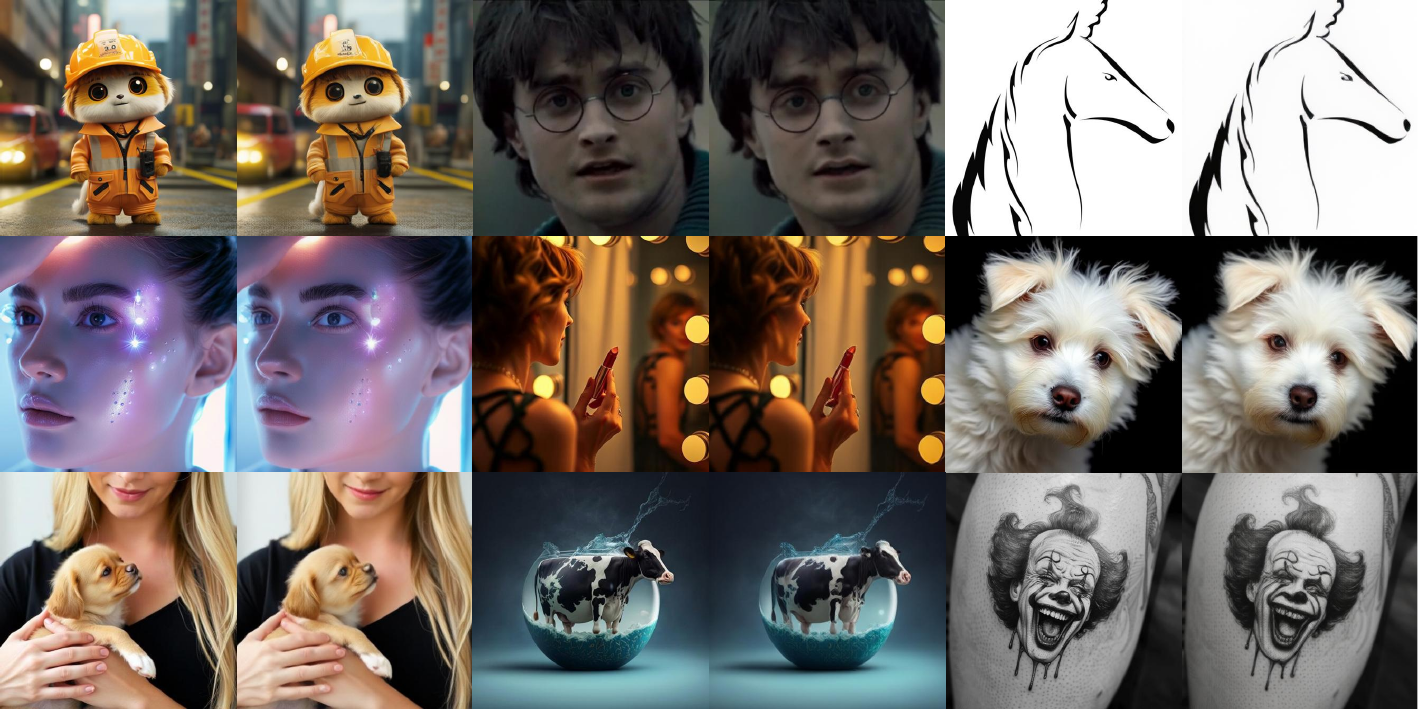}
    % \put(-6,33){(a)}
    % \put(-6,10){(b)}
    \end{overpic}
    \caption{
    Visualization of reconstruction results from {\methodname}-InternViT version.
    Left: input image; Right: output image.
    } 
    \label{fig:recon}
    \vspace{-18pt}
\end{figure}

As presented in Tab.~\ref{tab:tokenizer}, we evaluate reconstruction metrics on generative-only tokenizers and unified tokenizers. 
Our experiments demonstrate that employing pre-trained VFMs as unified encoders, ViT-based decoder, along with the discretization of semantic features, can achieve competitive reconstruction quality \textit{without any convolution blocks}. 
This finding aligns with the observation in~\cite{rae,unilip} that continuous features generated by a semantic encoder can be utilized for reconstruction tasks, while our work further validates the feasibility of operating in a \textbf{discrete} space.
In addition, our {\methodname} surpasses dual-encoder methods such as TokenFlow and MUSE-VL with a more efficient unified encoder style. We provide visualizations results in Fig.~\ref{fig:recon}.

\begin{table*}[t]
\centering
\scalebox{0.82}{
    \begin{tabular}{cccccccccccc}
    \toprule
    \textbf{Method} & \textbf{Vision Encoder} & \textbf{LLM} & \textbf{Res.} & \textbf{POPE} & \textbf{GQA} & \textbf{TQA} & \textbf{MMB} & \textbf{MME-P} & \textbf{SEED} & \textbf{MMMU} & \textbf{AI2D} \\
    \midrule
    \multicolumn{12}{c}{\textit{Understanding Only MLLM}} \\
    \midrule
    Emu3-Chat~\cite{emu3} & MoVQGAN & 8B from scratch & 512 & 85.2 & 60.3 & 64.7 & 58.5 & 1243.8 & 68.2 & 31.6 & 70.0 \\
    LLaVA-1.5\textsuperscript{\dag}~\cite{llava-1.5} & CLIP-L & Vicuna-7B & 336 & 85.9 & 62.0 & 46.1 & 64.3 & 1510.7 & 58.6 & 35.4 & 55.3 \\
    LLaVA-1.5\textsuperscript{\dag}~\cite{llava-1.5} & CLIP-L & Vicuna-13B & 336 & 85.9 & 63.3 & 61.3 & 67.7 & 1531.3 & 68.1 & 36.4 & 61.1 \\
    InternVL2.5~\cite{internvl2.5} & InternViT-300M & InternLM2.5-7B & 448 & 90.6 & - & 79.1 & 84.6 & - & - & 56.0 & 84.5 \\
    InternVL3~\cite{internvl3} & InternViT-300M & Qwen2.5-7B & 448 & 91.1 & - & 80.2 & 83.4 & 1748.4 & 77.1 & 62.7 & 85.2 \\
    Qwen2.5-VL~\cite{Qwen2.5-VL} & QwenViT & Qwen2.5-7B & 448 & 85.9 & - & 84.9 & 83.5 & 1698.1 & 77.0 & 58.6 & 83.9 \\
    \midrule
    \multicolumn{12}{c}{\textit{MLLM with Unified Tokenizer}} \\
    \midrule
    VILA-U\textsuperscript{\dag}~\cite{Vila-u} & SigLIP-so400m & Vicuna-7B & 256 & 81.6 & - & - & - & 1311.6 & - & - & - \\
    UniTok\textsuperscript{\dag}~\cite{Unitok} & Vitamin-L & Vicuna-7B & 256 & 81.7 & - & - & - & 1448.0 & - & - & - \\
    SemHiTok\textsuperscript{\dag}~\cite{Semhitok} & SigLIP-L & Vicuna-7B & 256 & 84.2 & 61.0 & - & 60.3 & 1400.6 & - & - & - \\
    QLIP\textsuperscript{\dag}~\cite{Qlip} & CLIP-L & Vicuna-7B & 392 & 86.1 & 61.8 & 55.2 & - & 1498.3 & - & - & - \\
    TokenFlow-L\textsuperscript{\dag}~\cite{tokenflow} & ViTamin-XL & Vicuna-13B & 256 & 85.0 & 60.3 & 54.1 & 60.3 & 1365.4 & 62.6 & 34.4 & 56.6 \\
    TokenFlow-XL~\cite{tokenflow} & SigLIP-so400m & Vicuna-13B & 384 & 86.8 & 62.7 & 61.5 & 68.9 & 1545.9 & 68.7 & 38.7 & 66.7 \\
    TokLIP\textsuperscript{\dag}~\cite{toklip} & ViT-so400m & Qwen2.5-7B & 384 & 82.7 & 59.3 & - & - & 1410.2 & 65.2 & 42.1 & - \\
    Tar~\cite{tar} & SigLIP2-so400m & Qwen2.5-7B & 384 & 87.8 & 61.3 & - & 74.4 & 1571.0 & 73.0 & 39.0 & - \\
    \rowcolor{cyan! 10} \textbf{{\methodname}}\textsuperscript{\dag} & SigLIP2-so400m & Vicuna-7B & 256 & 84.4 & 62.4 & 44.4 & 65.3 & 1445.7 & 66.4 & 31.3 & 53.1 \\
    \rowcolor{cyan! 10} \textbf{{\methodname}}\textsuperscript{\dag} & SigLIP2-so400m & Vicuna-13B & 256 & 85.1 & 63.4 & 46.5 & 65.5 & 1491.1 & 66.8 & 33.3 & 57.0 \\
    \rowcolor{cyan! 10} \textbf{{\methodname}}\textsuperscript{\dag} & SigLIP2-so400m & Vicuna-7B & 512 & 88.2 & 63.6 & 58.8 & 67.6 & 1494.2 & 62.8 & 33.9 & 55.3 \\
    \rowcolor{cyan! 10} \textbf{{\methodname}}\textsuperscript{\dag} & SigLIP2-so400m & Vicuna-13B & 512 & 88.2 & 64.8 & 61.7 & 67.3 & 1543.3 & 69.9 & 37.4 & 59.8 \\
    \rowcolor{cyan! 10} \textbf{{\methodname}} & InternViT-300M & Qwen2.5-7B & 448 & 90.5 & - & 80.6 & 85.1 & 1746.8 & 77.0 & 61.6 & 84.8 \\
    \bottomrule
    \end{tabular}
}

\caption{
Evaluation on multimodal understanding benchmarks.
We collect evaluations including: POPE~\cite{pope}; GQA~\cite{gqa}; TQA: TextVQA~\cite{textvqa}; MMB: MMBench-En~\cite{mmbench}; MME-P: MME-Perception~\cite{mme}; SEED: SEEDBench-Img~\cite{seedb}; MMMU~\cite{mmmu}; AI2D~\cite{ai2d}.
\dag\ denotes training on LLaVA-v1.5~\cite{llava-1.5} setting.
``Res.'' is an abbreviation of resolution.
Our pretrained {\methodname} can be directly comparable with SOTA open-sourced MLLMs without specific fine-tuning as detailed in Sec.~\ref{sec:und}.
}
\label{tab:understanding}
\vspace{-5pt}
\end{table*}

\subsection{Multimodal Understanding}
\label{sec:understanding}
As shown in Tab.~\ref{tab:understanding}, our {\methodname} consistently outperforms other unified tokenizers on downstream multimodal understanding benchmarks under the same training settings as LLaVA-1.5~\cite{llava-1.5}, both on 7B and 13B scales. 
It can be observed that the semantic representations provided by other unified tokenizers often cause information loss, manifested as varying degrees of performance degradation compared to the LLaVA-1.5 baseline~\cite{llava-1.5}. 
Moreover, our approach is more efficient, as it does not require multimodal alignment or instruction tuning for the pre-trained {\methodname} tokenizers (as presented in Sec.~\ref{sec:und}). 
By simply replacing the ViT encoder in the base model (\textit{e.g.}, InternVL3~\cite{internvl3}), our method can be directly applied to downstream tasks without performance degradation—in fact, performance may even improve. 
This suggests that our two-stage training approach introduced in Sec.~\ref{sec:unified} effectively preserves understanding performance while training for image reconstruction.
Our method yields a significant improvement on the MME-P~\cite{mme} benchmark compared to the dual-encoder method TokenFlow-L under the same setting of 13B parameters (1491.1 \textit{vs.} 1365.4). Also, we surpass Tar~\cite{tar} which performs semantic distillation directly on discrete tokens under the Qwen2.5-7B setting. It indicates the performance degradation on multimodal understanding due to quantization errors~\cite{vqgan,vqkd}. 

\subsection{Visual Generation}
\label{sec:genration}
\begin{table*}[t]
\centering
\scalebox{0.66}{
    \begin{tabular}{cc|ccccccc|cccccc}
    \toprule
    \multirow{2}{*}{\textbf{Method}} & \multirow{2}{*}{\textbf{\# Params}} & \multicolumn{7}{c}{\textbf{GenEval}~\cite{geneval}} & \multicolumn{6}{c}{\textbf{DPG-Bench}~\cite{dpgbench}} \\
    \cmidrule(lr){3-9} \cmidrule(lr){10-15} % 使用 cmidrule 来美化中间的分隔线
    & & \textbf{Single Obj.} & \textbf{Two Obj.} & \textbf{Counting} & \textbf{Colors} & \textbf{Position} & \textbf{Color Attri.} & \textbf{Overall$\uparrow$} & \textbf{Global} & \textbf{Entity} & \textbf{Attribute} & \textbf{Relation} & \textbf{Other} & \textbf{Overall$\uparrow$} \\
    \midrule
    \multicolumn{15}{c}{\textit{Diffusion-based Model}} \\
    \midrule
    SDv1.5~\cite{sd} & 0.9B & 0.97 & 0.38 & 0.35 & 0.76 & 0.04 & 0.06 & 0.43 & 74.63 & 74.23 & 75.39 & 73.49 & 67.81 & 63.18 \\
    PixArt-$\alpha$~\cite{pixart} & 0.6B & 0.98 & 0.50 & 0.44 & 0.80 & 0.08 & 0.07 & 0.48 & 74.97 & 79.32 & 78.60 & 82.57 & 76.96 & 71.11 \\
    SDv2.1~\cite{sd} & 0.9B & 0.98 & 0.51 & 0.44 & 0.85 & 0.07 & 0.17 & 0.50 & - & - & - & - & - & - \\
    SDXL~\cite{sdxl} & 2.6B & 0.98 & 0.74 & 0.39 & 0.85 & 0.15 & 0.23 & 0.55 & 83.27 & 82.43 & 80.91 & 86.76 & 80.41 & 74.65 \\
    DALLE3~\cite{dalle3} & - & 0.96 & 0.87 & 0.47 & 0.83 & 0.43 & 0.45 & 0.67 & 90.97 & 89.61 & 88.39 & 90.58 & 89.83 & 83.50 \\
    SD3-Medium~\cite{sd3} & 2B & 0.99 & 0.94 & 0.72 & 0.89 & 0.33 & 0.60 & 0.74 & 87.90 & 91.01 & 88.83 & 80.70 & 88.68 & 84.08 \\
    SANA-1.5~\cite{sana1.5} & 4.8B & 0.99 & 0.93 & 0.86 & 0.84 & 0.59 & 0.65 & 0.81 & - & - & - & - & - & 84.70 \\
    \midrule
    \multicolumn{15}{c}{\textit{Autoregressive-based Model}} \\
    \midrule
    Chameleon~\cite{chameleon} & 7B & - & - & - & - & - & - & 0.39 & - & - & - & - & - & - \\
    LlamaGen~\cite{llamagen} & 0.8B & 0.71 & 0.34 & 0.21 & 0.58 & 0.07 & 0.04 & 0.32 & 81.76 & 75.43 & 76.17 & 84.76 & 58.40 & 64.84 \\
    EMU3-Gen~\cite{emu3} & 8B & 0.98 & 0.71 & 0.34 & 0.81 & 0.17 & 0.21 & 0.54 & 85.21 & 86.68 & 86.84 & 90.22 & 83.15 & 80.60 \\
    TokenFlow~\cite{tokenflow} & 13B & 0.97 & 0.66 & 0.40 & 0.84 & 0.17 & 0.26 & 0.55 & 78.72 & 79.22 & 81.29 & 85.22 & 71.20 & 73.38 \\
    Janus~\cite{janus} & 1.3B & 0.97 & 0.68 & 0.30 & 0.84 & 0.46 & 0.42 & 0.61 & 82.33 & 87.38 & 87.70 & 85.46 & 86.41 & 79.68 \\
    SimpleAR~\cite{simplear} & 1.5B & - & 0.90 & - & - & 0.28 & 0.45 & 0.63 & 87.97 & - & - & 86.33 & - & 81.97 \\
    Janus-Pro~\cite{janus-pro} & 1B & 0.98 & 0.82 & 0.51 & 0.89 & 0.65 & 0.56 & 0.73 & 87.58 & 88.63 & 88.17 & 88.98 & 88.30 & 82.63 \\
    \rowcolor{cyan! 10} \textbf{{\methodname}} & 0.6B & 0.96 & 0.82 & 0.64 & 0.80 & 0.73 & 0.58 & 0.76 & 89.78 & 93.14 & 89.92 & 90.34 & 91.27 & 86.67 \\
    \bottomrule
    \end{tabular}
}

\caption{
Comparisons of visual generation quality on GenEval~\cite{geneval} and DPG-Bench~\cite{dpgbench}. Obj.: Object. Attri.: Attribute.
}
\label{tab:generation}
\end{table*}
We evaluate the performance of visual generation with our {\methodname} on the GenEval~\cite{geneval} and DPG-Bench~\cite{dpgbench} benchmarks as presented in Tab.~\ref{tab:generation}. 
Our lightweight visual generation models with only 0.6B parameters showcase competitive capabilities with models of similar parameter size.
It indicates that semantic high dimensional latent space built on VFMs not only accelerates the convergence of diffusion-based models for generative modeling~\cite{repa,rae}, but also benefits the training dynamics of the autoregressive models in discrete scaling paradigm.

\subsection{Ablation Studies}
\label{sec:ablation}

\textbf{Codebook Dim.} 
Previous studies~\cite{llamagen,vqgan,ibq} found that when quantizing features extracted from CNN-based encoders to learn a VQ codebook, a relatively small dimension (ranging from 8 to 256) should be used—a choice that has been interpreted as necessary for reconstruction requiring more fine-grained details. 
These approaches often encounter issues such as codebook collapse or a sharp decline in utilization when further scaling up to the typical dimensions of CLIP-based encoders (e.g., 1152). 
However, as shown in Tab.~\ref{tab:abl_codebook}, it presents a contrary conclusion: when quantizing features extracted from VFMs (ViT-based), the dimension of the codebook should be higher; otherwise, it leads to training non-convergence and codebook collapse. 
In particular, our {\methodname} maintains a high utilization ratio (nearly 100\%) in a high dimension range.

\hspace{-0.4cm}\textbf{Codebook Size.}
We also ablate the effect of codebook size on our unified tokenizer {\methodname} in Tab.~\ref{tab:abl_codebook}.
We observe that the reconstruction quality continually improves with increasing codebook size.
However, when codebook size exceeds 16K, we observe a slight degradation which is attributed to the slow convergence of the training process.

% \input{tabs/abl_VFMs}

% \hspace{-0.4cm}\textbf{Choice of VFMs.}
% Similar to RAE~\cite{rae}, we conduct ablation experiments on the choice of different VFMs for further exploration. 
% As presented in Tab.~\ref{tab:abl_vfms}, we find that DINOv2~\cite{dino} and InternViT~\cite{internvl3} showcase competitive reconstruction quality. 
% However, we select SigLIP2~\cite{siglip2} and InternViT~\cite{internvl3} to conduct our main experiments for better alignment with pretrained LLMs.

\begin{table}[h]
\centering
\resizebox{0.95\linewidth}{!}{
    \begin{tabular}{cc|cccc}
    \toprule
    \textbf{Dim} & \textbf{Size} & \textbf{rFID$\downarrow$} & \textbf{PSNR$\uparrow$} & \textbf{SSIM$\uparrow$} & {\textbf{Ratio}$\uparrow$} \\
    \midrule
    $\leq 256$ & \multirow{6}{*}{16384} & NA & NA & NA & NA \\
    384 & & 7.69 & 8.24 & 0.261 & 64\% \\
    768 & & 5.38 & 13.76 & 0.398 & 69\% \\
    1152 & & 3.51 & 17.22 & 0.569 & 83\% \\
    \cellcolor{cyan! 10} \textbf{1536} & & \textbf{2.65} & \textbf{20.14} & \textbf{0.668} & \textbf{100\%} \\
    1920 & & 2.69 & 20.07 & 0.664 & 98\% \\
    \midrule
    \multirow{4}{*}{1536} & 4096 & 7.07 & 8.02 & 0.253 & 100\% \\
    & 8192 & 3.74 & 17.02 & 0.548 & 100\% \\
    & \cellcolor{cyan! 10} \textbf{16384} & \textbf{2.65} & \textbf{20.14} & \textbf{0.668} & \textbf{100\%} \\
    & 32768 & 2.78 & 19.94 & 0.645 & 96\% \\
    \bottomrule
    \end{tabular}
}
\caption{
Ablation on the hyperparameters of the VQ codebook.
}
\vspace{-15pt}
\label{tab:abl_codebook}
\end{table}
\begin{table}[h]
\centering
\resizebox{1.0\linewidth}{!}{
    \begin{tabular}{cc|ccc|cccc}
    \toprule
    \multirow{2}{*}{\textbf{Two}} & \multirow{2}{*}{\textbf{Self-}} & \multicolumn{3}{c}{\textbf{Reconstruction}} & \multicolumn{4}{c}{\textbf{Understanding}} \\
    \cmidrule(lr){3-5} \cmidrule(lr){6-9}
    \textbf{Stage} & \textbf{Distillation} & \textbf{rFID$\downarrow$} & \textbf{PSNR$\uparrow$} & \textbf{SSIM$\uparrow$} & \textbf{MME-P$\uparrow$} & \textbf{MMB$\uparrow$} & \textbf{AI2D$\uparrow$} & \textbf{TQA$\uparrow$} \\
    \midrule
    \ding{55} & \ding{55} & \textbf{2.69} & \textbf{21.35} & \textbf{0.704} & 608.9 & 22.3 & 48.6 & 7.0 \\
    \ding{55} & \ding{51} & 2.84 & 19.68 & 0.644 & 1435.2 & 64.9 & 52.8 & 42.6 \\
    \rowcolor{cyan! 10} \ding{51} & \ding{51} & 2.71 & 20.52 & 0.680 & 1439.1 & \textbf{65.8} & \textbf{53.1} & \textbf{44.0} \\
    \bottomrule
    \end{tabular}
}

\caption{
Ablation on the effect of training strategy on image understanding and reconstruction.
}
\vspace{-24pt}
\label{tab:abl_training}
\end{table}
\begin{figure}[h]
    \centering
    \begin{overpic}[width=1.0\linewidth]{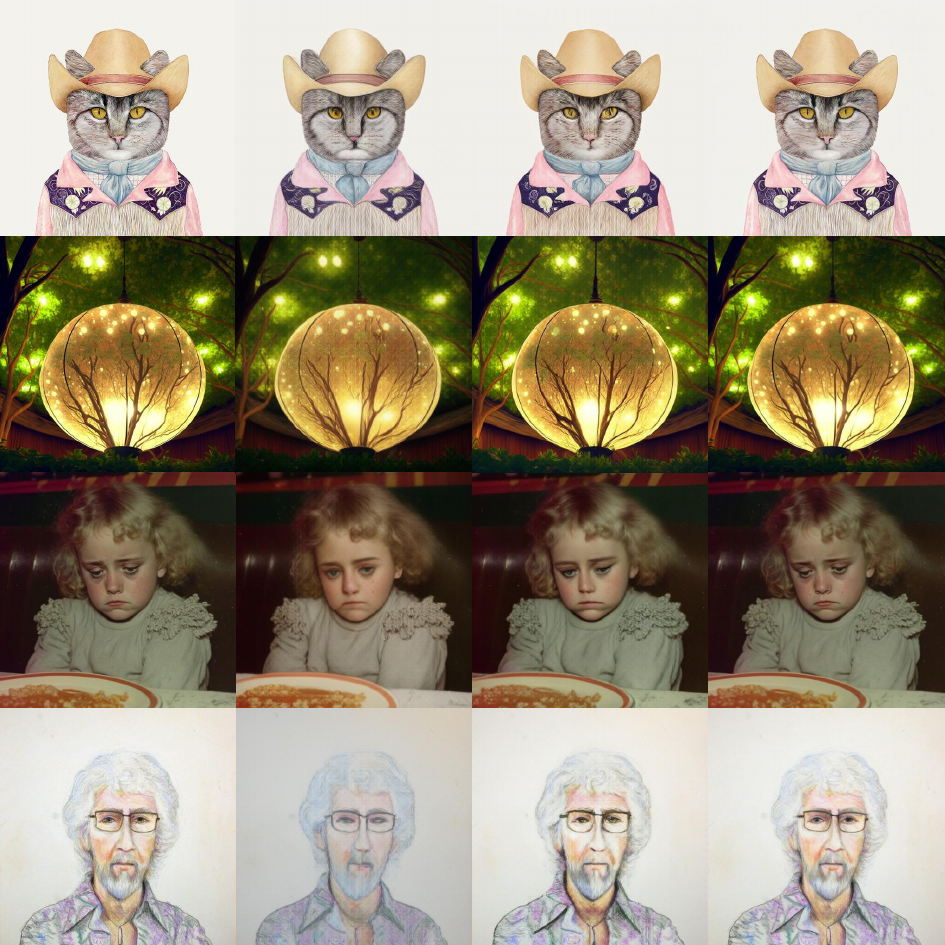}
    \put(5,102){Original}
    \put(32,102){Stage 1}
    \put(57,102){Stage 2}
    \put(85,102){E2E}
    % \put(-6,10){(b)}
    \end{overpic}
    \caption{
    Visualization results on ablation study of training strategies. 
    As indicated in Tab.~\ref{tab:abl_training}, the second training stage adds more fine-grained details on reconstruction and retains semantics, while end-to-end training without distillation constraints fails to achieve a trade-off between them.
    }
    \label{fig:rec_abl}
    \vspace{-15pt}
\end{figure}
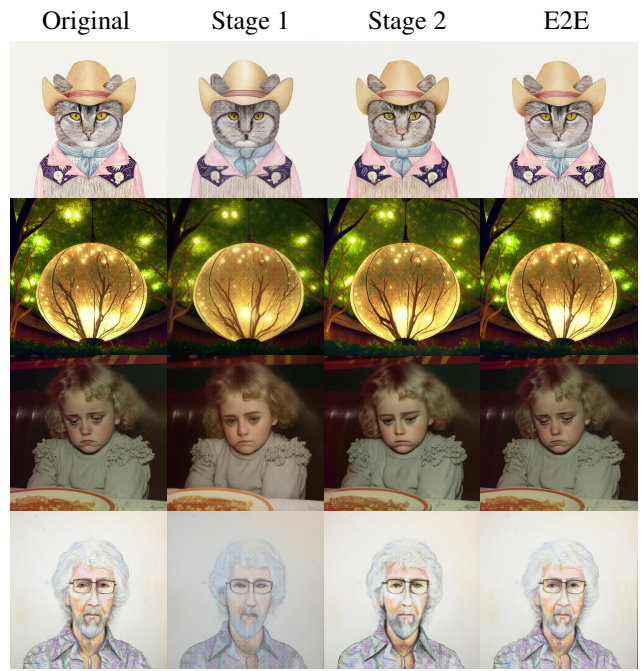

\hspace{-0.4cm}\textbf{Training Strategies.}
We conduct ablation experiments on the training strategy of unifying understanding, generation and reconstruction in a single tokenizer in Tab.~\ref{tab:abl_training}. 
The first row represents end-to-end training all the components without self-distillation, which achieves the best reconstruction quality and fails to maintain the performance of visual understanding. 
The second row utilizes semantic constraints and alleviates the degradation of the ability of visual understanding.
It is advisable to minimize the update of well-pretrained encoder in the initial training for image reconstruction.
As shown in Fig.~\ref{fig:rec_abl}, with the two-stage training strategy and self-distillation loss, we achieve a trade-off on image reconstruction and understanding.

\section{Conclusion}
\label{sec:conclusion}
In this paper, we propose {\methodname}, a vector quantization variant of RAE designed for unified tokenizers, which pioneers the first attempt to produce continuous semantic representations for multimodal understanding and fine-grained discrete tokens for visual generation.
We eliminate the dependency on pixel encoders during the training of the unified tokenizer, employing a pure ViT-based model and a two-stage training strategy to achieve the integration of visual understanding, generation, and reconstruction. 
Built upon pre-trained VFMs, {\methodname} is the first to realize high-utilization, high-dimensional codebooks for discrete autoregressive modeling. 
Extensive experiments on the multimodal understanding, generation and reconstruction benchmarks demonstrate the competitive performance compared to both diffusion-based generative models and autoregressive models with unified tokenizers.

\section{Acknowledgements}

This work was supported by Kuaishou Technology. We extend our sincere gratitude to all collaborators involved in this project. Moreover, this work is also supported by the National Key R\&D Program of China (2022YFB4701400/4701402), SSTIC Grant (KJZD20230923115106012, KJZD20230923114916032, GJHZ20240218113604008).

\newpage

{
    \small
    \bibliographystyle{ieeenat_fullname}
    \bibliography{main}
}

\clearpage
\appendix
\setcounter{page}{1}
\maketitlesupplementary

In the supplementary materials, Section~\ref{appdx:overview} provides a concise overview of the motivation underlying this study and highlights the distinctions between {\methodname} and other related works; 
Section~\ref{appdx:impl} illustrates the detailed training configurations of the tokenizer (Section~\ref{appdx:tok_train}) and autoregressive models for understanding and generation (Section~\ref{appdx:ar_train}) and their evaluation setups (Section~\ref{appdx:eval}).
Section~\ref{appdx:qual_res} presents more qualitative results in image reconstruction (Section~\ref{appdx:rec_res}), visual generation (Section~\ref{appdx:gen_res}) and failure cases (Section~\ref{appdx:failure}).
Finally, we summarize the limitations of our work and future exploration in Section~\ref{appdx:limit_and_future}.

\section{Overview}
\label{appdx:overview}

\textbf{\textit{Related Work.}} Reviewing the evolution of unified models, pioneering works such as Chameleon~\cite{chameleon} and EMU-3~\cite{emu3} employed VQGAN~\cite{vqgan} as the encoder for both understanding and generation tasks, albeit at the expense of semantic understanding. The Janus series~\cite{janus,janus-pro,janusflow} adopted a dual-encoder architecture, utilizing a semantic encoder like CLIP~\cite{clip} for multimodal understanding and a pixel encoder such as VQGAN~\cite{vqgan} for image generation. This approach, however, hindered interaction and alignment between the two types of representations. The unified tokenizer aims to employ a single encoder-decoder framework to maintain performance on understanding tasks while enabling generation. Nevertheless, many previous methods~\cite{tokenflow,toklip,Dualtoken,DualViTok,muse-vl,Qlip,Vila-u,Semhitok,Unitok} were overly complex in design or relied on simple contrastive learning to provide weak semantic supervision on latent tokens, making it difficult to achieve a satisfactory trade-off between comprehension and reconstruction. These tokenizers generally still underperform compared to classic understanding-only baseline models like LLaVA-1.5~\cite{llava-1.5}. Tar~\cite{tar} and X-Omni~\cite{x-omni} were among the first works to utilize pretrained Visual Foundation Models (VFMs) as encoders while discretizing representations to preserve multimodal understanding capabilities. Although they narrowed the performance gap of earlier discrete tokenizers~\cite{chameleon,emu3} on understanding tasks, they still suffer from quantization errors and lack inherent autoencoder properties. Recent work, RAE~\cite{rae}, discovered that high-dimensional ViT encoders can be directly applied to reconstruction and have been used to replace original VAEs~\cite{vae} in diffusion-based generative models~\cite{dit}.

\hspace{-0.4cm}\textbf{\textit{Contributions.}} The contribution of our paper can be summarized in three folds: (i) We propose a vector quantization version of RAE, namely \textbf{{\methodname}}, which pioneers the first attempt in unifying understanding, generation and reconstruction. (ii) {\methodname} is the first unified tokenizer to produce continuous semantic features for understanding and fine-grained discrete tokens for generation and reconstruction \textbf{simultaneously}. (iii) {\methodname} features the first \textbf{high-dimensional} VQ codebook (comparable to CLIP encoders) with a nearly \textbf{100\%} utilization ratio, which is contrary to previous explorations in this field.

\section{Implementation Details}
\label{appdx:impl}

\subsection{Tokenizer Training Details}
\label{appdx:tok_train}
{\methodname} is pretrained on BLIP3-o~\cite{blip3-o} open-sourced data, which consists of 27M samples recaptioned by Qwen2.5-VL-7B~\cite{Qwen2.5-VL}, 5M samples from CC12M~\cite{cc12m}, and 4M synthesized images from JourneyDB~\cite{journeydb} (Table~\ref{tab:pilot} \& ~\ref{tab:tokenizer}). We train three variants of encoder: InternViT-300M-448px~\cite{internvl3}, SigLIP2-so400m-256px~\cite{siglip2} and SigLIP2-so400m-512px~\cite{siglip2}. The decoder adopts the symmetric design with encoder. Specific training configurations are provided in Table~\ref{tab:appdx_tok_training}. The code implementation is adapted from TiTok~\cite{titok} repository.

\begin{table*}[t]
\centering
\scalebox{0.62}{
    \begin{tabular}{c|ccc|ccc}
    \toprule
    & \multicolumn{3}{c}{\textbf{Stage1}} & \multicolumn{3}{c}{\textbf{Stage2}} \\
    \midrule
    \textbf{Model} & \textbf{SigLIP2-so400m} & \textbf{SigLIP2-so400m} & \textbf{InternViT-300M} & \textbf{SigLIP2-so400m} & \textbf{SigLIP2-so400m} & \textbf{InternViT-300M} \\
    \cmidrule(lr){1-1} \cmidrule(lr){2-4} \cmidrule(lr){5-7}
    resolution & 256px & 512px & 448px & 256px & 512px & 448px \\
    freeze encoder & true & true & true & false & false & false \\
    codebook size & 16384 & 16384 & 16384 & 16384 & 16384 & 16384 \\
    codebook dim & 1536 & 1536 & 1536 & 1536 & 1536 & 1536 \\
    discriminator start steps & NA & NA & NA & 50000 & 50000 & 30000 \\
    discriminator weight & NA & NA & NA & 0.1 & 0.1 & 0.1 \\
    distillation weight & NA & NA & NA & 1.0 & 1.0 & 1.0 \\
    perceptual loss weight & 1.1 & 1.1 & 1.1 & 1.1 & 1.1 & 1.1 \\
    perceptual model & convnext-s & convnext-s & convnext-s & convnext-s & convnext-s & convnext-s \\
    augmentation & random crop \& random flip & random crop \& random flip & random crop \& random flip & random crop \& random flip & random crop \& random flip & random crop \& random flip \\
    encoder lr & NA & NA & NA & 1e-5 & 1e-5 & 1e-5 \\
    decoder lr & 4e-4 & 4e-4 & 4e-4 & 1e-4 & 1e-4 & 1e-4 \\
    end lr & 1e-4 & 1e-4 & 1e-4 & 1e-5 & 1e-5 & 1e-5 \\
    scheduler & cosine & cosine & cosine & cosine & cosine & cosine \\
    weight decay & 1e-4 & 1e-4 & 1e-4 & 1e-4 & 1e-4 & 1e-4 \\
    discriminator lr & NA & NA & NA & 4e-5 & 4e-5 & 4e-5 \\
    optimizer & AdamW & AdamW & AdamW & AdamW & AdamW & AdamW \\
    ($\beta_1, \beta_2$) & (0.9, 0.999) & (0.9, 0.999) & (0.9, 0.999) & (0.9, 0.999) & (0.9, 0.999) & (0.9, 0.999) \\
    warmup steps & 2000 & 2000 & 2000 & 2000 & 2000 & 2000 \\
    mixed precision & bf16 & bf16 & bf16 & bf16 & bf16 & bf16 \\
    max grad norm & 1.0 & 1.0 & 1.0 & 1.0 & 1.0 & 1.0 \\
    global batch size & 1024 & 1024 & 768 & 1024 & 1024 & 768 \\
    total steps & 100000 & 100000 & 45000 & 70000 & 70000 & 60000 \\
    \bottomrule
    \end{tabular}
}

\caption{
The detailed training configurations of {\methodname} tokenizer.
}
% \vspace{-24pt}
\label{tab:appdx_tok_training}
\end{table*}

\subsection{Autoregressive Training Details}
\label{appdx:ar_train}
For image understanding, we follow the setups in LLaVA-1.5~\cite{llava-1.5}, which uses LLaVA-Pretrain-595K and LLaVA-v1.5-mix-665K for pretraining and SFT (Table~\ref{tab:pilot} \& ~\ref{tab:understanding} \& ~\ref{tab:abl_training}). For visual generation, we train it on BLIP3-o~\cite{blip3-o} data and additional 80M high quality images (Table~\ref{tab:generation}). 
We conduct ablation studies on VQ codebook on ImageNet-1K with 20 epochs training for efficiency (Table~\ref{tab:abl_codebook} \& \ref{tab:abl_training}). Note that, the results in Table~\ref{tab:abl_training} are attained through training on stage 1. The LLM backbones include: Vicuna-v1.5-7B~\cite{vicuna}, Vicuna-v1.5-13B~\cite{vicuna} and Qwen2.5-7B~\cite{qwen2.5} for understanding; Qwen3-0.6B~\cite{qwen3} for generation. \textbf{Besides, we did not conduct specific training on understanding tasks for our pretrained tokenizer.} Specific training configurations are provided in Table~\ref{tab:appdx_und_training} \& ~\ref{tab:appdx_gen_training}. The code implementation is adapted from the LLaVA~\cite{llava-1.5} repository.

\begin{table}[h]
\centering
\resizebox{1.0\linewidth}{!}{
    \begin{tabular}{c|cc}
    \toprule
    \textbf{Hyperparameters} & \textbf{Pretrain} & \textbf{SFT} \\
    \midrule
    freeze backbone & true & false \\
    freeze ViT & true & true \\
    freeze connector & false & false \\
    mm\_projector\_type & mlp2x\_gelu & mlp2x\_gelu \\
    mm\_vision\_select\_layer & -1 & -1 \\
    deepspeed stage & ZeRO-2 & ZeRO-3 \\
    scheduler & cosine & cosine \\
    learning rate & 1e-3 & 2e-5 \\
    weight decay & 0.0 & 0.0 \\
    optimizer & AdamW & AdamW \\
    ($\beta_1, \beta_2$) & (0.9, 0.999) & (0.9, 0.999) \\
    warmup ratio & 0.03 & 0.03 \\
    mixed precision & bf16 & bf16 \\
    max grad norm & 1.0 & 1.0 \\
    global batch size & 256 & 128 \\
    model max length & 2048 & 2048 \\
    \bottomrule
    \end{tabular}
}

\caption{
The detailed training configurations of multimodal understanding, SigLIP2-Vicuna-v1.5-7B / 13B.
}
% \vspace{-24pt}
\label{tab:appdx_und_training}
\end{table}
\begin{table}[h]
\centering
\resizebox{1.0\linewidth}{!}{
    \begin{tabular}{c|c}
    \toprule
    \textbf{Hyperparameters} & \textbf{Generation Tuning} \\
    \midrule
    augmentation & center crop \\
    deepspeed stage & ZeRO-1 \\
    scheduler & constant \\
    learning rate & 1e-4 \\
    weight decay & 0.0 \\
    optimizer & AdamW \\
    ($\beta_1, \beta_2, \epsilon$) & (0.9, 0.999, 1e-15) \\
    warmup steps & 4000 \\
    mixed precision & bf16 \\
    max grad norm & 1.0 \\
    global batch size & 512 \\
    model max length & 1536 \\
    \bottomrule
    \end{tabular}
}

\caption{
The detailed training configurations of visual generation.
}
% \vspace{-24pt}
\label{tab:appdx_gen_training}
\end{table}

\subsection{Evaluation Details}
\label{appdx:eval}
For image reconstruction, we evaluate on the 256 x 256 ImageNet 50k validation set to compute the rFID, PSNR and SSIM as in TiTok~\cite{titok} official codebase in Table~\ref{tab:pilot} \& ~\ref{tab:tokenizer}. For multimodal understanding, we evaluate the SigLIP2-Vicuna variants using the LMMs-Eval codebase~\cite{zhang2024lmmsevalrealitycheckevaluation,lmms_eval2024}. We directly replace the ViT with our tokenizer without specific training. Due to the compatibility with InternVL3~\cite{internvl3}, we utilize the OpenCompass VLMEvalKit~\cite{duan2024vlmevalkit} for evaluation of InternViT-Qwen2.5-7B in Table~\ref{tab:understanding}. For visual generation, we use the official evaluation toolkit from GenEval~\cite{geneval} and DPG-Bench~\cite{dpgbench} in Table~\ref{tab:generation}.

\section{Additional Qualitative Results}
\label{appdx:qual_res}
We provide more qualitative results on image reconstruction and generation in this section. Also, we present some failure cases in our tokenizer and AR model.

\subsection{Reconstruction Results}
\label{appdx:rec_res}
\begin{figure*}[h]
    \centering
    \begin{overpic}[width=1.0\linewidth]{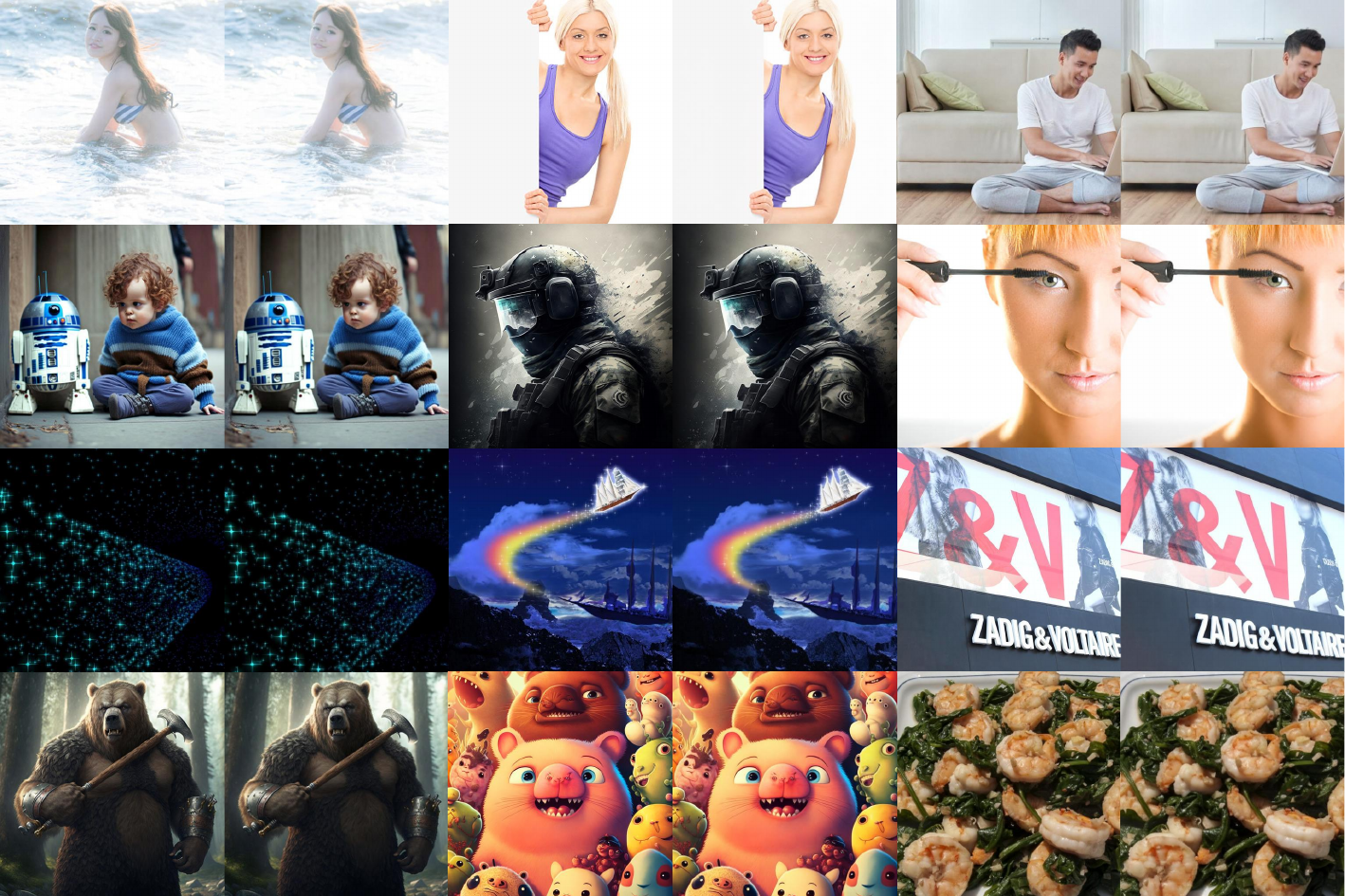}
    % \put(-6,33){(a)}
    % \put(-6,10){(b)}
    \end{overpic}
    \caption{
    Additional visualization of reconstruction results from {\methodname}-InternViT.
    Left: input image; Right: output image.
    } 
    \label{appdx:rec}
    % \vspace{-18pt}
\end{figure*}

As shown in Figure~\ref{appdx:rec}, our {\methodname} can achieve fine-grained reconstruction in human faces, scenes and objects.

\subsection{Generation Results}
\label{appdx:gen_res}
\begin{figure*}[h]
    \centering
    \begin{overpic}[width=1.0\linewidth]{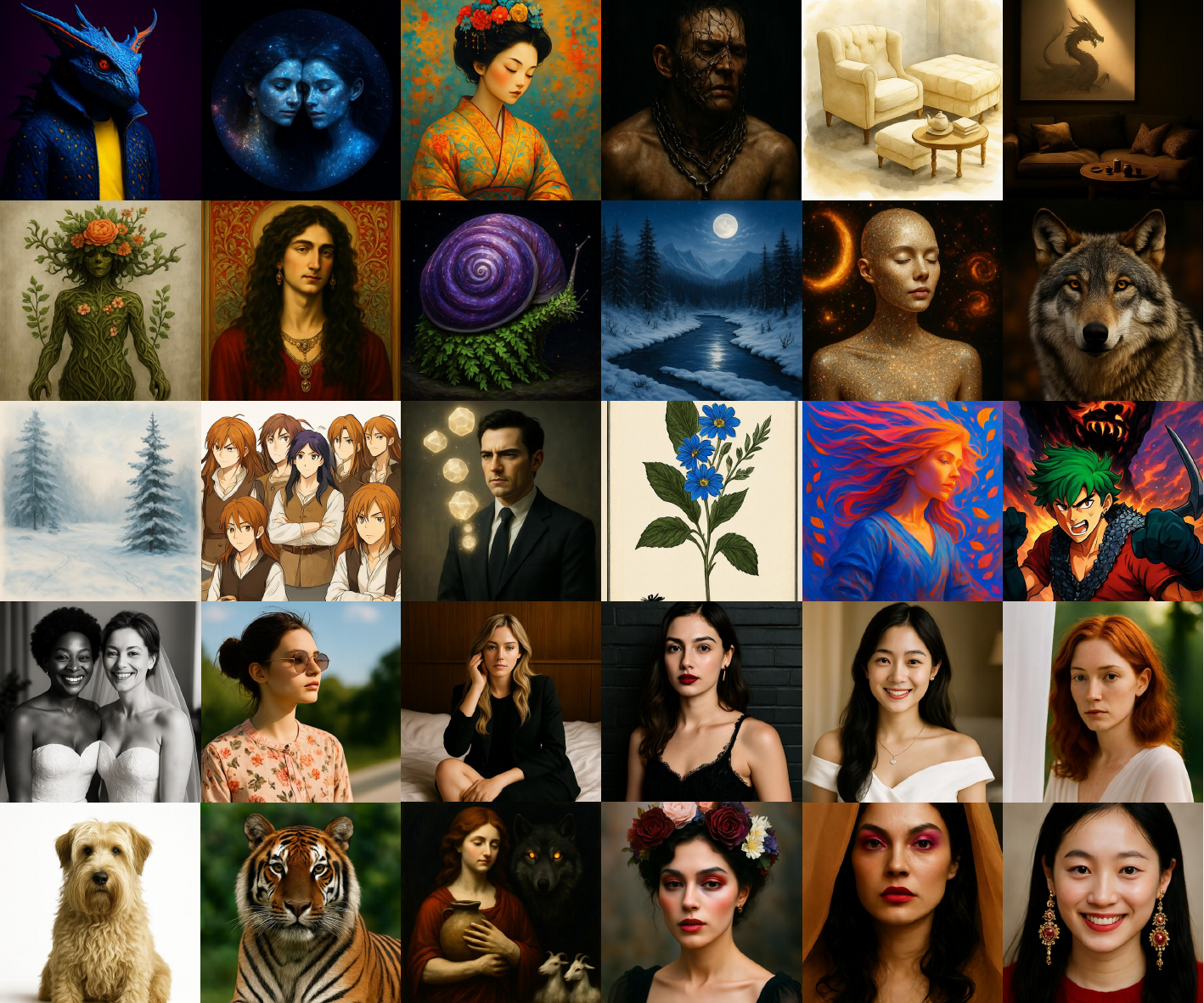}
    % \put(-6,33){(a)}
    % \put(-6,10){(b)}
    \end{overpic}
    \caption{
    Additional visualization of generation results at 512 x 512 px.
    } 
    \label{appdx:gen}
    % \vspace{-18pt}
\end{figure*}

We present additional visual generation results in Figure~\ref{appdx:gen}. Our method can generate images with various styles, subjects, and scenarios.

\subsection{Failure Cases}
\label{appdx:failure}
\begin{figure*}[h]
    \centering
    \begin{overpic}[width=0.78\linewidth]{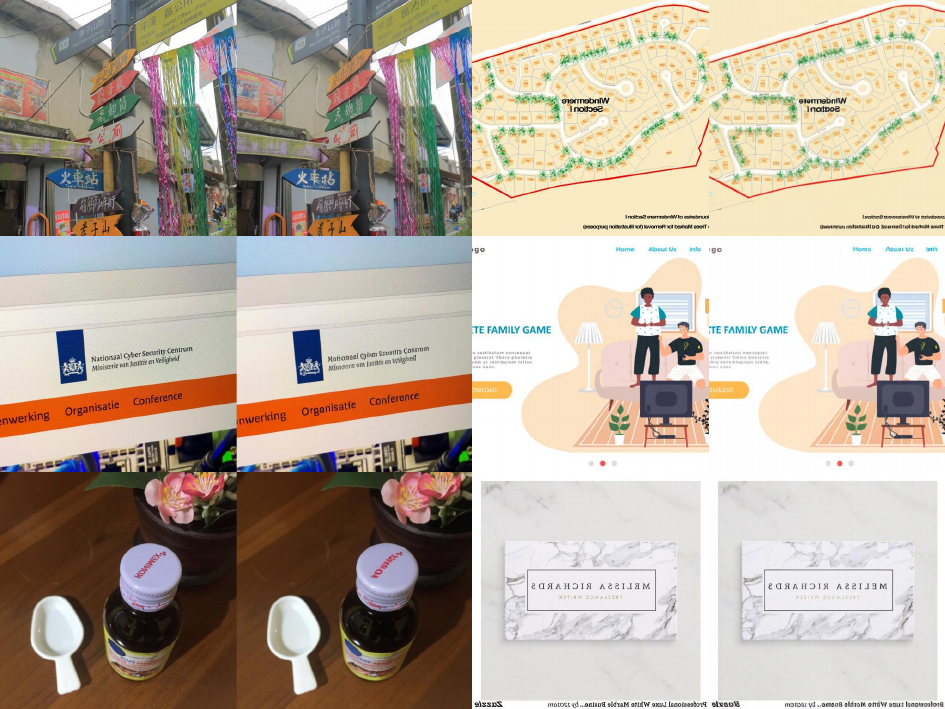}
    % \put(-6,33){(a)}
    % \put(-6,10){(b)}
    \end{overpic}
    \caption{
    Failure reconstruction cases from {\methodname}-InternViT.
    Left: input image; Right: output image.
    } 
    \label{appdx:failure_tok}
    % \vspace{-18pt}
\end{figure*}

\begin{figure*}[h]
    \centering
    \begin{overpic}[width=0.78\linewidth]{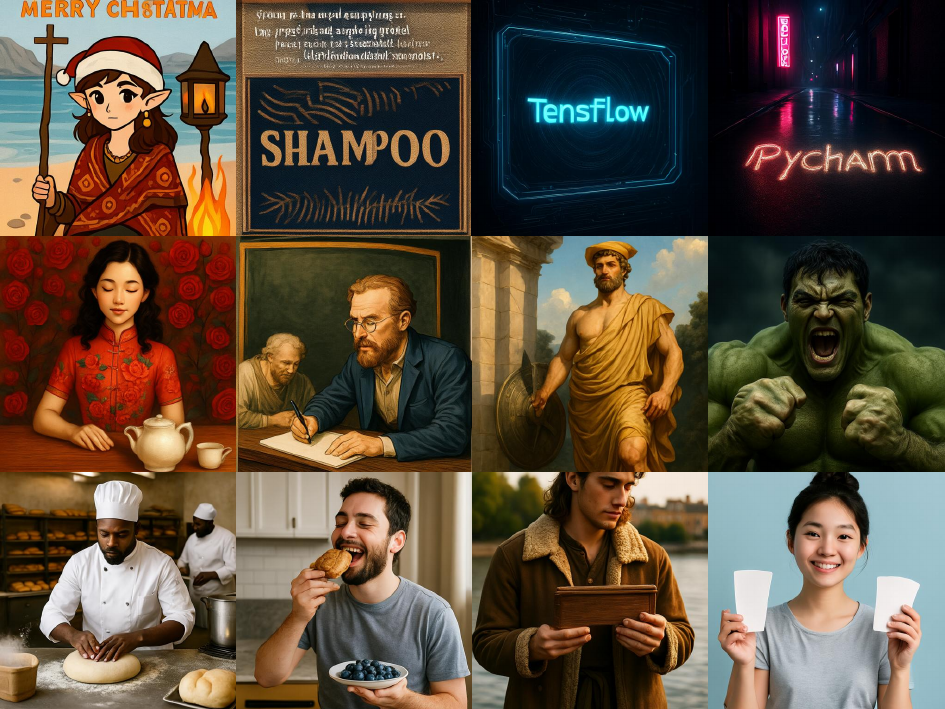}
    % \put(-6,33){(a)}
    % \put(-6,10){(b)}
    \end{overpic}
    \caption{
    Failure generation cases. Our model still has certain artifacts in generating fine-grained text, small human faces and fingers, which can be addressed with extensive training data and reinforcement learning as explored in~\cite{x-omni,simplear,emu3.5}.
    } 
    \label{appdx:failure_gen}
    % \vspace{-18pt}
\end{figure*}

As shown in Figure~\ref{appdx:failure_tok}, our tokenizer remains flawed in text reconstruction and high-density scenarios, which is likely attributable to the trade-off between semantic representation and reconstruction performance and specific text data tuning. Moreover, in terms of image generation, in Figure~\ref{appdx:failure_gen}, certain artifacts persist in fingers and human faces, issues that may primarily necessitate resolution through post-training~\cite{x-omni,blip3-o,simplear}.

\section{Limitation and Future Work}
\label{appdx:limit_and_future}
The primary limitation of {\methodname} lies in the lack of exploration of alternative and more effective methods to balance understanding and reconstruction performance to minimize the compromise on understanding capability. The potential for reconstruction and generation to enhance understanding remains underexplored. Additionally, the quantization loss inherent in the discrete tokenizer makes it challenging for {\methodname} to compete with state-of-the-art continuous VAEs. There is still room for improvement in generation quality, particularly in understanding spatial relationships, texture rendering, and addressing artifacts in faces and fingers.

Looking ahead, several directions~\cite{x-omni,emu3,Nextstep-1,chartbench,chartmoe,chartpoint,alore,wei2025learning,wei2025unifying,uniglyph,chunkgrpo} appear particularly promising for future research. In this work, our main objective is to develop a unified tokenizer that provides more effective representations for understanding, generation, and reconstruction tasks. However, leveraging such representations to integrate various tasks into a single model requires further investigation. Issues such as conflicts and synergies among different tasks, as well as efficient model scaling, are left for future work.



\end{document}